\documentclass[10pt,twocolumn,letterpaper]{article}

\usepackage[pagenumbers]{iccv} 

\usepackage{graphicx}
\usepackage{booktabs}
\usepackage{nicefrac}       
\usepackage{microtype}      
\usepackage{xcolor}         
\usepackage[dvipsnames]{xcolor}
\usepackage{colortbl}
\usepackage{amsmath}
\usepackage{amssymb}
\usepackage{mathtools}
\usepackage{amsthm}
\usepackage{amsmath}
\usepackage{caption}
\usepackage{subcaption}
\usepackage{enumitem}
\usepackage{makecell}
\usepackage{float}
\usepackage{placeins}
\usepackage{mathtools}
\usepackage{amsthm}
\usepackage{lineno}
\usepackage{epsfig}
\usepackage{comment}
\usepackage{multirow}
\usepackage{adjustbox}
\usepackage{textcomp}
\usepackage{wrapfig}
\usepackage{setspace}
\usepackage{kantlipsum, wrapfig,graphicx}
\usepackage[rightcaption]{sidecap}
\usepackage{caption}
\usepackage{soul}
\usepackage[ruled,linesnumbered,noend,vlined]{algorithm2e}
\usepackage{algorithmicx,algpseudocode}
\usepackage{float}
\usepackage{listings}
\usepackage{fancybox}
\usepackage{tcolorbox}
\tcbuselibrary{skins}

\usepackage{algpseudocode}
\usepackage{multirow}

\definecolor{iccvblue}{rgb}{0.21,0.49,0.74}
\usepackage[pagebackref,breaklinks,colorlinks,allcolors=iccvblue]{hyperref}

\def\paperID{14} 
\def\confName{ICCV}
\def\confYear{2025}

\title{REVEAL: Robust Evolution of Vision-Language Models \\ for Explainable AI-Video Detection}

\author{~~Yun-Yun Tsai$^1$,~~Qingyuan Liu$^1$,~~Ruijian Zha$^1$,~~Victoria Li$^1$,\\
~~Pengyuan Shi$^1$,~~Chengzhi Mao$^2$,~~Junfeng Yang$^1$\\ \\
Columbia University\\
{\tt\small $^1$\{yunyuntsai,junfeng\}@cs.columbia.edu $^2$\{cm1838\}@scarletmail.rutgers.edu }\\
{\tt\small $^1$\{ql2505,rz2689,vl2395,ps3391\}@columbia.edu} \\
}

\newcommand{\foo}[3][\linewidth]{%
   \par\noindent
   \sbox0{\texttt{#2:}\hspace{1mm}#3}%
   \ifdim\wd0<#1%
      \usebox0%
   \else
      \sbox0{\texttt{#2:}}%
         \copy0
         \hspace{1mm}
         \parbox[t]{\dimexpr#1-1em-\wd0-2\fboxsep-2\fboxrule\relax}{#3}%
   \fi
}

\lstdefinestyle{base}{
  emptylines=1,
  breaklines=true,
  moredelim=**[is][\color{blue}]{@}{@},
}

\newcommand{\reveal}[0]{REVEAL\xspace}
\newcommand{\vidfor}[0]{VidForensic\xspace}

\begin{document}
\maketitle
\begin{abstract}
The rapid advancement of AI-generated video poses challenges to digital authenticity and security. Current detection methods, often trained on specific datasets, struggle with the ever-evolving landscape of generative techniques and unseen manipulations. We introduce a framework leveraging Vision Language Models (VLMs) for robust AI-generated video detection. Our approach equips the VLM with the ability to reason about video content and use external tools to identify subtle inconsistencies, mirroring human system 2 thinking. Our self-evolving VLM dynamically selects and composes appropriate tools, enhancing its ability to generalize to novel video generation techniques. The modular design promotes interpretability, allowing for a clearer understanding of VLM's decision-making process. To evaluate, we establish the first benchmark \vidfor containing 1.4k+ high-quality AI generated videos across eight generative models. Experiments show that \reveal improves F1 scores by 9.1\% to 30.2\% over top baselines across our datasets for VLMs, notably for GPT-4o, Gemini 1.5 pro, and QWen-VL-Max, and Llava-One-Vision-7B. While open-world AI-video detection remains an open challenge, our results indicate that existing methods fail primarily because they lack tool-enabled, higher-order reasoning.

\end{abstract}

\section{Introduction}

\begin{figure}[t!]
    \centering
    \includegraphics[width=.999\linewidth]{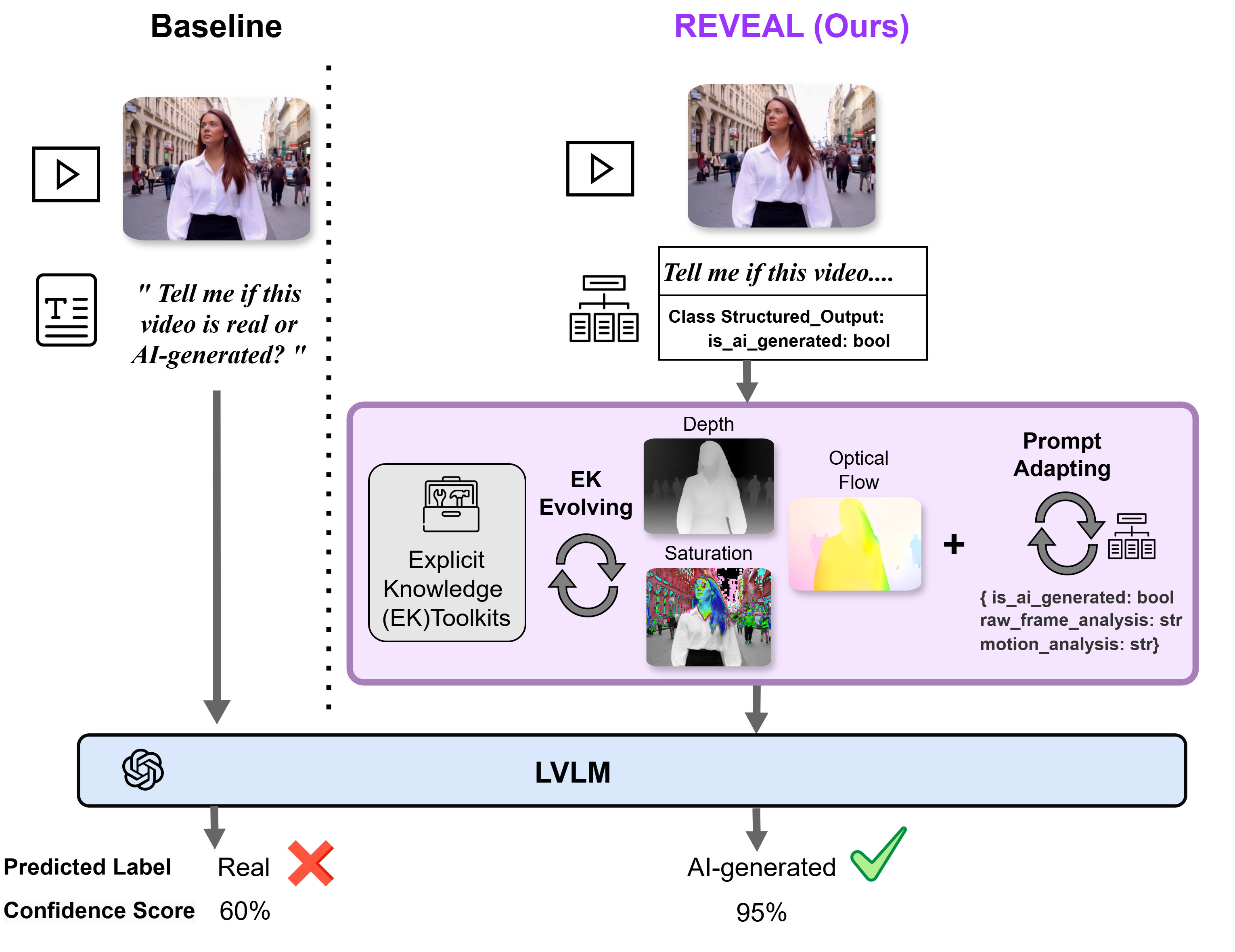}
    \caption{An example of an AI-generated video from Kling~\cite{klingai} where \reveal makes a correct prediction with explicit knowledge selection and prompt self-adaptation. \reveal facilitates VLMs for video detection by calling explicit knowledge tools to extract useful information from the original videos and by providing structure-formatted output.}
    \label{fig:main_flow}
\end{figure}

\label{sec:intro}
The realm of video creation is undergoing a significant transformation with the advent of video generation tools, such as Stable Video Diffusion~\cite{blattmann2023stable}, SORA by OpenAI~\cite{videoworldsimulators2024}, Runway Gen3~\cite{runway3}, Pika~\cite{pika}, and Show-1~\cite{zhang2023show}. These cutting-edge tools are revolutionizing industries, ranging from design and marketing to entertainment and education, by creating high-quality video content. While this pivotal shift is opening up a myriad of possibilities for creators everywhere, it also poses societal dangers, particularly in its widespread use for spreading disinformation, propaganda, scams, and phishing -- evidenced by cases such as the Taylor Swift deepfakes~\cite{taylorswift}. These potential threats underscore the importance of detecting videos created by these generative models.

Prior work often extract artifacts from collected samples and train auxiliary deep neural networks to act as detectors~\cite{chang2024matters}. However, these supervised learning-based methods face fundamental generalization issues due to the high dimensionality of video data, because collecting exhaustive data in high-dimensional space is impossible. Moreover, the increasingly high quality video generated from current diffusion models~\cite{wang2019cnngenerated, patchforeccv20} makes it increasingly harder to detect. Furthermore, even powerful VLMs, despite their strong contextual understanding, struggle to identify key inconsistencies in AI-generated videos when using simple, free-form prompts, highlighting the need for more structured reasoning and targeted information extraction.

\begin{figure*}[t]
    \centering
    \includegraphics[width=0.8\textwidth]{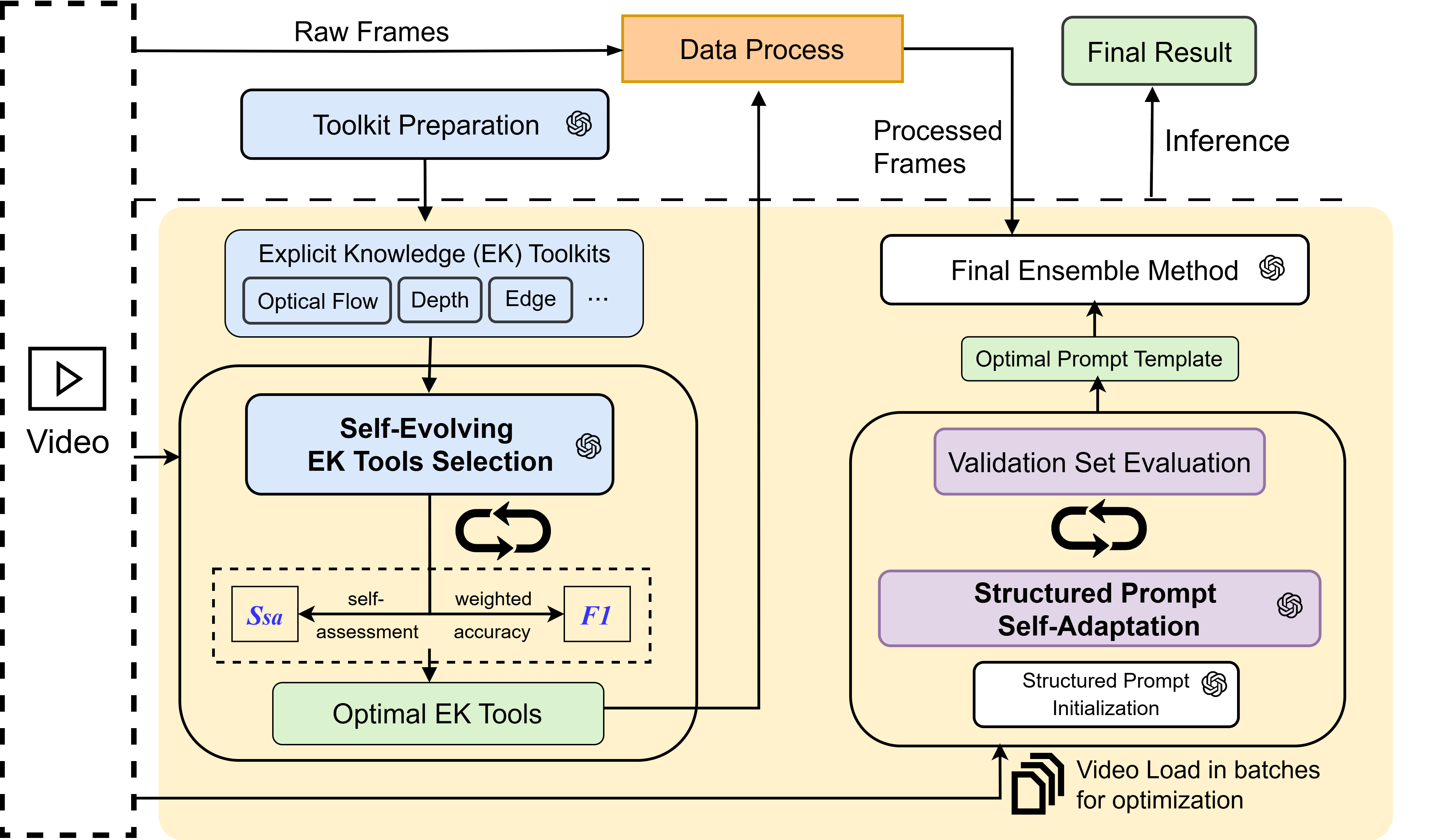}
    \caption{We show the main pipeline of \textbf{\reveal} for video detection. First, VLMs suggest relevant video detection tools. Next, based on the weighted accuracy, $\text{F1}$, and the self-assessment score, $S_{SA}$, each tool provides, we assemble a customized toolkit for each video detection VLM. 
   In Fig.~\ref{fig:selfadapt-prompt}, we show details of the self-adaptation of the structured prompt. The prompt tuning will be based on the VLM itself. Component marked with the logo$~\includegraphics[width=3mm, height=2.5mm]{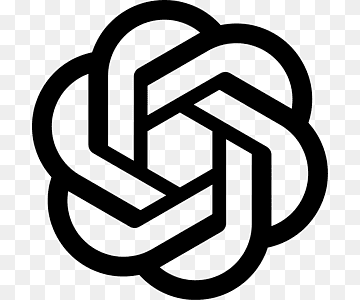}$ are developed with the VLM-like GPT-4o~\cite{openai2024gpt4o}.}
    \label{fig:main_flow}
\end{figure*}


We introduce \emph{\reveal}, a self-evolving framework for AI-generated video detection using VLMs. Unlike methods relying solely on direct visual feature extraction, REVEAL empowers the VLM to actively reason about video content and dynamically select and compose a toolkit of external analysis tools (e.g., for motion analysis, artifact detection). As illustrated in Fig.~\ref{fig:main_flow}, this process unfolds in two key steps: (1) \textbf{Self-Evolving Toolkit Selection}, where the VLM, acting as a judge, autonomously evaluates and selects the most effective tools based on predefined scoring metrics and historical samples; and (2) \textbf{Self-Adaptation with Structured Prompting}, where the VLM refines the structure of its prompts based on output feedback, enhancing adaptability across diverse datasets. This data-driven, self-evolving approach, guided by structured prompts with a defined class structure, allows \reveal to automatically learn optimal reasoning strategies and tool combinations, improving interpretability, reducing hallucination, and preventing overfitting through online adaptation of key prompt fields.

A key advantage of our method is interpretability, which stems from its modular design and tool-extracted explicit knowledge, revealing the decision-making process. The self-evolving capabilities, using structured prompting and dynamic tool selection, achieve significant robustness and generalizability. REVEAL's adaptability makes it less vulnerable to overfitting and more capable of handling novel generation techniques, a crucial step towards reliable open-world detection.

We highlight our main contributions: 
\begin{itemize}
    \item  We present a novel method that enables VLMs to perform AI-generated video detection tasks precisely through an automated self-evolving framework, consisting of two main steps: (1) self-evolving to select EK Toolkit; (2) self-adaptation with structured prompt tuning.
    \item Our pipeline is training-free and only requires a lightweight evolving process for updating model preferences of EK selection and adjusting prompt templates.  
    \item We create a new benchmark \vidfor, including 1.4k+ high-quality fake videos generated from multiple sources of video generation tools, such as Kling~\cite{klingai}, Runway Gen3~\cite{runway3}, and OpenSORA~\cite{opensora}. To facilitate future research, we also provide the nine corresponding EKs extracted from each video frame in \vidfor.
    \item Evaluation results show that \reveal improves F1 scores by 9.1\% to 30.2\% over the direct prompt baselines on high-quality datasets across four state-of-the-art(SOTA) VLMs: Qwen-VL-Max~\cite{qwen2023qwenvl}, Gemini-1.5-pro~\cite{google2024gemini}, GPT-4o~\cite{openai2024gpt4o}, and Llava-OV-7B~\cite{li2024llavaonevisioneasyvisualtask}. \reveal also surpasses the conventional supervised learning video detection methods.
\end{itemize}

\section{Related Works}
\label{sec:relatedworks}


\paragraph{\textbf{AI-Generated Video Detection}}
The success in high-quality machine-generated videos has heightened concerns about security, personal privacy, and digital integrity, emphasizing the need for a robust and generalizable detector capable of distinguishing videos produced by generative models.
Recently, Deepfake video, generated by GAN-based models, can perform face manipulation with high realism~\cite{deepfake_pavel}. Agarwal et al.~\cite{deepfake_pavel} highlighted the challenges of detecting Deepfake video, noting that traditional DNN networks or audio-visual approach based on lip sync inconsistency detection are not robust enough to detect Deepfake. David et al.,~\cite{8639163_deepfake} proposed to use convolutional neural network (CNN) + Long short-term memory (LSTM) to build Deepfake video detectors. However, these methods did not account for cross-model transferability to state-of-the-art synthetic videos, especially those generated by diffusion-based video models~\cite{opensora,blattmann2023stablevideodiffusionscaling,klingai,runway3,pika}, which have the capability to produce high-quality synthetic videos indistinguishable by human. VidProM~\cite{wang2024vidprommillionscalerealpromptgallery} and DeMamba~\cite{chen2024demambaaigeneratedvideodetection} addressed the challenge and created synthetic video datasets containing millions of samples. DIVID~\cite{liu2024turnsimrealrobust} further refined the diffusion reconstruction error (DIRE~\cite{wang2023dire}) for diffusion-generated video detection, enhancing detection performance across temporal dimensions. AIGVDet~\cite{bai2024aigeneratedvideodetectionspatiotemporal} proposed to use spatio-temporal CNNs to tackle synthetic video detection.  DuB3D~\cite{ji2024distinguishfakevideosunleashing} developed dual-branch 3D transformers to distinguish between real and synthetic videos. Despite the inspiring in-domain evaluation results from previous work, the robustness and generalizability of existing detectors' performance on unseen sources remain unexplored.

\paragraph{\textbf{Video Detection with LVLM}}

Large Vision-Language Models (LVLMs) have emerged as a powerful framework for integrating visual and textual data, enabling models to perform complex multimodal tasks. Early LVLMs, such as CLIP~\cite{radford2021learningtransferablevisualmodels}and ALIGN~\cite{jia2021scalingvisualvisionlanguagerepresentation}, excel at mapping images and text into a shared embedding space, enabling efficient image recognition and task captioning. However, these models are limited in their ability to understand temporal information in video data. To address this, models like Flamingo~\cite{alayrac2022flamingovisuallanguagemodel} and MERLOT~\cite{zellers2021merlotmultimodalneuralscript} have been introduced, significantly advancing LVLM capabilities in video understanding. Additionally, BLIP-2~\cite{li2023blip2bootstrappinglanguageimagepretraining} improved LVLM performance in image understanding by refining multimodal fusion techniques, enhancing the model's ability to comprehend nuanced relationships between visual objects and their linguistic descriptions. These models have paved the way for applying LVLMs to complex multimodal applications such as Video Question Answering and Image Understanding.




\section{Preliminary}
\label{preliminary}
\subsection{Task Definition}
Our objective is to explore the reasoning capabilities of LVLMs in detecting videos generated by any source of video generative models. Given a video input $v$ and a corresponding selected set of explicit knowledge ($\text{EK}$), we ask LVLM to classify $v$ as candidate label $y=\{\textit{real}, \textit{AI}\}$ based on following criteria: (1) Whether there are artifacts from the selected set ($\text{EK}$) for $v$. (2) Whether there are inconsistencies from the selected set ($\text{EK}$).  Here, we view each tool in $\text{EK}$ as an individual detection sub-task. 

\subsection{Video Dataset Exploration}

To facilitate our research, we create a new benchmark called \emph{\textbf{\vidfor}}. \vidfor dataset features 200 text-to-video prompts and more than 1.4k high-quality videos, collected or generated from eight generative models. In Table~\ref{tab:video_dataset_detail}, we show the details of \vidfor benchmark.
For real videos, we collect them from PANDA-70M~\cite{chen2024panda70mcaptioning70mvideos}, a real-world video dataset with millions of videos sourced from YouTube. 
For fake videos, we either collect them from VidProM~\cite{wang2024vidprommillionscalerealpromptgallery} or generate them by ourselves to incorporate the latest generative models. 
To ensure video quality, during the VidProM collection process, we carefully filter out low-quality videos (e.g., with background inconsistencies, subject inconsistencies, or unsmooth motion) by using \emph{VBench}\footnote{VBench, video quality assessment tool. \url{https://github.com/Vchitect/VBench}.}, a SOTA video quality assessment tool. 
For our in-house video dataset, we utilize SOTA generation tools, such as OpenSORA, Kling~\cite{klingai}, and Runway Gen3~\cite{runway3}, to generate high-quality videos based on the 200 prompts collected from the captions in PANDA-70M videos. In Appendix~\ref{supp:prompt_generation_process}, we provide details of the high-quality prompt generation process.
Unlike previous AI-generated video benchmarks DeMamba~\cite{chen2024demambaaigeneratedvideodetection} and DuB3D ~\cite{distinguisgFakeDataset}, \vidfor contains not only video raw samples but also the nine corresponding EKs that can facilitate future research use.


\begin{table}[t]
\centering
\scriptsize
\setlength{\tabcolsep}{1pt} 
\begin{tabular}{l|l|ccccc}
\toprule
 \textbf{Dataset Source} &\textbf{Video Source} & \textbf{Type} & \textbf{\# Videos}  & \textbf{Res.} & \textbf{FPS} & \textbf{Length} \\
\midrule
PANDA-70M~\cite{chen2024panda70mcaptioning70mvideos} & Youtube & Real & 200 & - & - & 1$\sim$10s \\
\midrule
VidProM~\cite{wang2024vidprommillionscalerealpromptgallery} & Text2Video-Zero~\cite{khachatryan2023text2videozerotexttoimagediffusionmodels} & AI & 200 & 512*512 & 4 & 2s \\
 & VideoCrafter2~\cite{chen2024videocrafter2overcomingdatalimitations} & AI & 200 & 512*320 & 10 & 1s \\
 & ModelScope~\cite{wang2023modelscopetexttovideotechnicalreport} & AI & 200 & 256*256 & 8 & 2s \\
 & Pika~\cite{pika} & AI & 200 & - & 24 & 3s \\

\midrule

Self-Collected & Youtube & Real & 45  & - & 30 & 1$\sim$4s \\
 & SORA~\cite{videoworldsimulators2024} & AI & 45 & - & 30 & 8$\sim$60s \\

\midrule

Self-Generated & OpenSORA~\cite{opensora} & AI & 200  & 1280*720 & 24 & 4s \\
 & Kling~\cite{klingai} & AI & 200 & 1280*720 & 30 & 5s \\
 & Runway-Gen3~\cite{runway3} & AI & 200 & 1280*768 & 30 & 5$\sim$10s \\

\bottomrule
\end{tabular}
\vspace{-1mm}
\caption{Composition of the \vidfor. We collect high-quality video from multiple sources. For dataset source own-generated, we generate text-to-video samples with generators conditioned on text prompts collected from PANDA-70M~\cite{chen2024panda70mcaptioning70mvideos} by ourselves.}
\label{tab:video_dataset_detail}
\vspace{-1mm}
\end{table}

\begin{table}[t]
\centering
\scriptsize
\setlength{\tabcolsep}{2pt} 
\begin{tabular}{l|l}
\toprule
\textbf{Category}   & \textbf{Explicit Knowledge (EK) Toolkits}                     \\ \midrule
\textbf{Appearance} & Saturation, Denoised,  Sharpen, Enhance, Segmentation Map \\
\textbf{Motion}     & Optical flow, Landmark                                        \\
\textbf{Geometry}   & Depth map, Edge                                               \\ \bottomrule
\end{tabular}
\vspace{-1mm}
\caption{Categories of explicit knowledge toolkits. Though all tools are proposed by LVLMs, we list and categorize all explicit knowledge  that we collect from LVLM in the process of initial toolkit preparation into three VR categories.}
\label{tab:ek_category}
\vspace{-2mm}
\end{table}

\subsection{Explicit Knowledge Exploration}
Recent research has shown that explicit knowledge extracted from video samples can help improve detection on video forensic~\cite{chang2024mattersaigeneratedvideos}. The explicit knowledge is collected from the video representations (VR) decomposed by the video frames. The VRs can be categorized into three classes~\cite{chang2024mattersdetectingaigeneratedvideos}: \textit{appearance}, \textit{motion}, and \textit{geometry}. Appearance refers to the visual attributes of video frames, such as color, lightning, or texture, while motion refers to the temporal or dynamic change in video frames, such as optical flow. Finally, geometry refers to object shape structure and spatial information in video frames, such as 3D depth map. We explore LVLM's comprehension capabilities from these three VR aspects. Our pipeline leverages LVLM to automatically select a set of explicit knowledge that can benefit the detection performance.
In Table~\ref{tab:ek_category}, we categorize EK toolkits into three VR classes.
In Appendix~\ref{supp:explicit-knowleged-intro}, we provide details of each EK.

\section{Method}
\label{sec:method}

\subsection{Self-Evolving LVLM with Explicit Knowledge}
We introduce how \reveal self-evolves LVLM with explicit knowledge (EK) enhancement and selection.
One of the key findings in our work is that different LVLMs show various capabilities in utilizing EK features. For example, GPT-4o has better knowledge on saturation and can offer a more reasonable explanation when compared to other LVLMs such as Gemini~\cite{google2024gemini} or Qwen~\cite{qwen2023qwenvl}. Therefore, to achieve optimal EK utilization, we design a model-specific EK selection strategy to construct a toolkit for each model.  For each model, we dynamically assign the best performing tools, selected from the entire EK pool, to its toolkit based on its performance on a small set of reference video samples. 
\paragraph{Model-Specific EK Selection Strategy} To explore reasoning abilities of models objectively and subjectively, we leverage both weighted accuracy and self-assessment score. Specifically, given a tool set $t_i \in \text{EK}\{t_1, ..., t_q\}$ and a subset of reference samples $x \in \mathcal{X}$, where $q=9$ is the number of tools, our designed tool-selection metrics $S_{\text{Tool}}$ compute scores for each tool $t$ upon model $\mathcal{M}$, taking both subjective evaluation and objective weighted accuracy of the model into consideration. We describe the score as follows:

\begin{equation}
    S_{\text{Tool}}(t, x) = \alpha \cdot \text{F1}_{\text{weighted}}(t,x) + (1 - \alpha) \cdot S_{\text{SA}}(t) \ \nonumber
\label{eqn:selection_metrics}
\end{equation}
where the $\text{F1}_{\text{weighted}}(\cdot)$ is the confidence-weighted F1 score, reflecting an objective view of the model, given tool $t$ for samples $x \in \mathcal{X}$. 
Specifically, given $N$ samples, each sample $x_i$ has $y_i \in \{real, AI \}$ as ground truth. 
We process $x_i$ with given tool $t$ and extract the explicit knowledge feature $z_i$. The
model's prediction is $\mathcal{M}(z_i)=\hat{y}_i \in \{real, AI\}$ and confidence score is $c_i \in [0, 1]$. We calculate $\text{F1}_{\text{weighted}}$ with weighted $\text{TP}$, $\text{FP}$, and $\text{FN}$. For instance, the weighted true positive ($\text{TP}$) is denoted as $\sum_{i=1}^N c_i \cdot 1 (y_i=real, \hat{y}_i=real)$, where $1(\cdot)$ is an indicator function. The self-assessment score $S_{\text{SA}}(\cdot)$ reflects the subjective view of models on the given tool $t$. An example message is provided below for prompting the LVLM to give us $S_{\text{SA}}$ for tool $t$ based on self-assessment.
\vspace{-2mm}

\begin{tcolorbox}[colback=white!5!white,colframe=gray!2!gray,arc=2mm,fontupper=\small]
\begin{adjustbox}{minipage=[c][0.60\linewidth]{1\linewidth}}
- Prompts: "You are given an AI-generated video detection task. Assess the the additional feature: \{\texttt{tool name}\} that could support your determination. \\
- Analysis History: \{\texttt{current fewshot results}\}

Evaluate your own analysis considering these factors:
\vspace{-1mm}
 \begin{tcolorbox}[colback=blue!20!white,,colframe=white!5!white, top=0.5pt,bottom=0.5pt]
\begin{lstlisting}[basicstyle=\footnotesize]
* Alignment with knowledge base
* Interpretability and transparency
* Robustness across scenarios
\end{lstlisting}
\end{tcolorbox}
\vspace{-1mm}
- Scoring: Provide a score from 0 to 10 based on your self-assessment. Higher score indicates an effective feature.
\end{adjustbox}
\end{tcolorbox}

$\alpha$ is a weighting factor that balances the relative importance of  F1 score against other evaluation metrics. 
We set $\alpha$ as 0.5. 

\paragraph{Tool selection by thresholding}
After calculating $S_{\text{Tool}}$ for each tool $t_i \in \text{EK}\{t_1, ..., t_q\}$, we select tools from EK pool for model $\mathcal{M}$ based on a baseline threshold. We define the threshold as
\begin{equation}
    S_{\text{Baseline}}(x) = \alpha \cdot \text{F1}_{\text{weighted}}(x) + (1 - \alpha) \cdot S_{\text{SA}}(t=\texttt{"RGB"}), \ \nonumber
\label{eqn:selection_metrics}
\end{equation}
where F1 score is calculated with raw samples $x \in \mathcal{X}$ and $S_{\text{SA}}$ is calculated by giving $\{$\texttt{tool name}$\}$ as \texttt{"RGB"}.
 The optimal set $\text{EK}^\star$ is composed by $t_i \in \text{EK}\{t_1, ..., t_q\}$ with a smaller $S_{\text{Tool}}$ compared to $S_{\text{Baseline}}$.
\begin{equation}
   t_i \in \text{EK}\{t_1, ..., t_q\} =
   \begin{cases}
      1, & \text{if}\ S_{\text{Tool}(t_i)} \geq S_{\text{Baseline}(t_i)} \\
      0, & \text{otherwise}
    \end{cases}  \ \nonumber
\label{eqn:selection_metrics}
\end{equation}

\subsection{Self-Adaptation with Structured Prompts}

In this paper, we explore two prompting approaches, free-formatted prompting and structured prompting. 
We directly prompt the LVLM with the message, formatted according to the template shown in Fig.~\ref{fig:prompt_exp}.
\begin{figure}[t]
\begin{tcolorbox}[colback=white!5!white,colframe=gray!2!gray,arc=2mm,fontupper=\small]
\begin{adjustbox}{minipage=[c][0.36\linewidth]{1\linewidth}}

\foo{role}{System}
\foo{content}{You are an AI video analyzer. Determine if
            the video is AI-generated or not?}
\foo{role}{User}
\foo{content}{Video: $\{$
                        "\textit{text-decription}": "\textit{These 8 images are consecutive frames of a video.}", "\textit{image-url}": [\textit{url}]$\}$}
\vspace{1mm}
\foo{Result}{$\{\text
{Default}\}$ or $\{\textcolor{blue}{\text{Structured Response}}\}$}
\end{adjustbox}
\end{tcolorbox}
\vspace{-2mm}
\begin{tcolorbox}[colback=white!5!white,colframe=gray!2!gray,arc=2mm,fontupper=\small]
\begin{adjustbox}{minipage=[c][0.20\linewidth]{1\linewidth}}
\begin{lstlisting}[basicstyle=\footnotesize, style=base]
class @Structured_Response@(BaseModel):
    is_ai_generated: bool
    raw_frame_analysis: str
    {tool_name}_analysis: str
    explanation: str
\end{lstlisting}
\end{adjustbox}
\end{tcolorbox}
\vspace{-5mm}
\caption{Prompt example for LVLM}
\label{fig:prompt_exp}
\end{figure}

Recent works~\cite{,zhong2022proqastructuralpromptbasedpretraining} on pre-trained LVLM indicated that the advantages of structured response may have tight connection among QA reasoning tasks, ranging from diverse question types, domains, to answer types. A structurally designed prompt-based input schema can help model the knowledge commonalities for general detection tasks while keeping knowledge customization for different EK~\cite{openaistructured}.
Motivated by this, we leverage structured prompt on several representative models for detection.

\begin{figure}[t]
    \centering
    \includegraphics[width=0.49\textwidth]{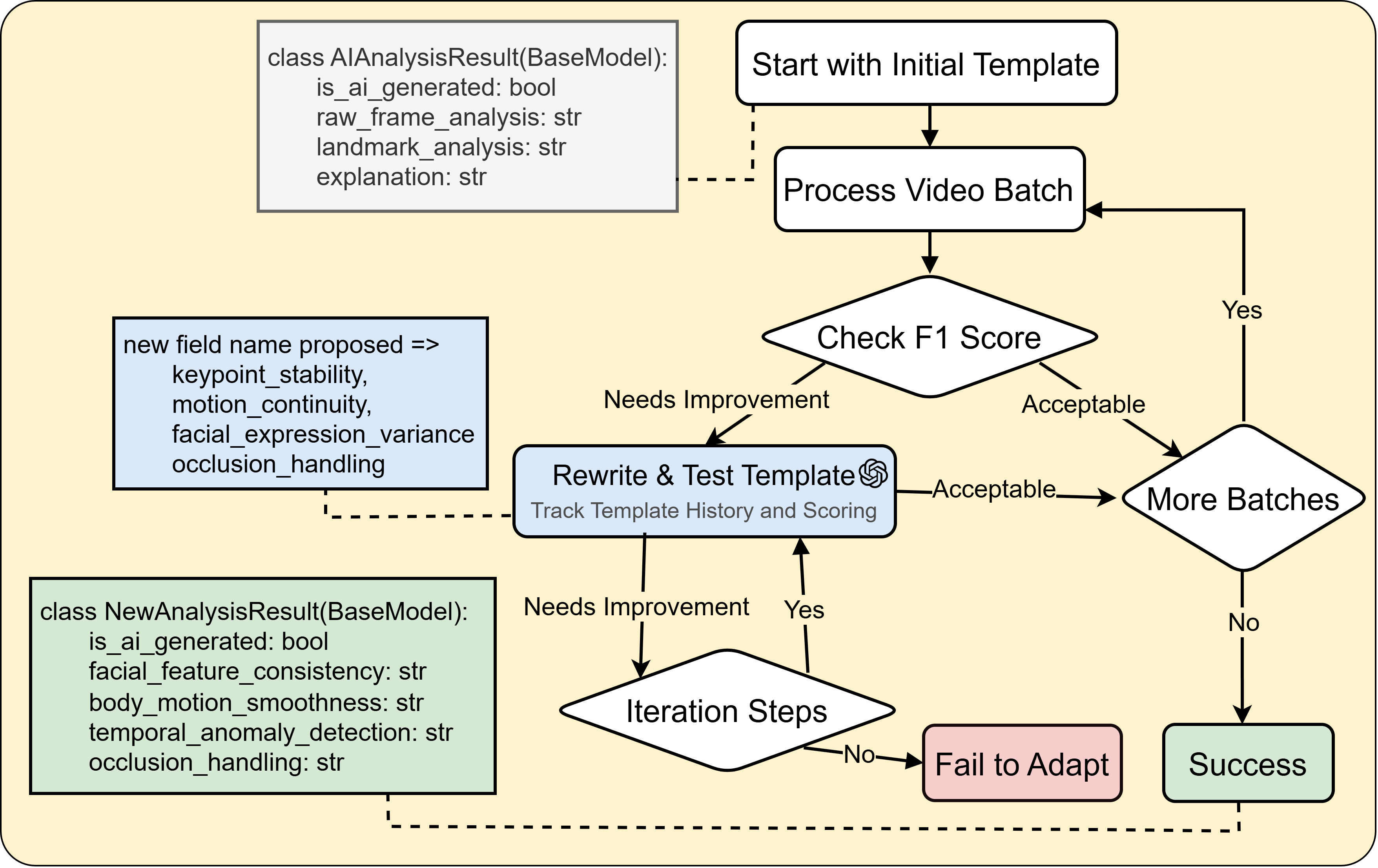}
    \caption{The flow of self-adaptation prompting process in \textbf{\reveal}. Component marked with the logo$~\includegraphics[width=3mm, height=2.5mm]{figs/gpt_logo.png}$ are developed with the LVLM like GPT-4o~\cite{openai2024gpt4o}.}
    \label{fig:selfadapt-prompt}
\end{figure}

\paragraph{Self-Adaptation Prompting Strategy} We dynamically re-write the structured prompt template of LVLM based on real-time performance feedback. 
In Fig.~\ref{fig:selfadapt-prompt}, we show the flow of self-adaptation prompting process. Our template re-write function is a key component of the process. Given small batches of video input, this function ensures that the LVLMs adapt their structured prompts to specific analytical needs. A detailed system prompt consists of five key criteria: \textit{Requirements, Analysis Guidelines, Constraints, Prohibited Fields,} and \textit{Additional Notes}. We ask the LVLMs to follow these \textit{five} criteria and modify the prompts. For instance, in the \textit{Requirements}, we have the rule said \textit{“You may add any other fields you deem necessary to analyze video data with} \texttt{tool name}\textit{.”}; in \textit{Analysis Guidelines"}, we have \textit{“Consider the aspects of videos that} \texttt{tool name} \textit{excels at analyzing.”}; and in the \textit{Additional Notes}, we have \textit{”The total number of fields must not exceed five”}...,etc. More information can be found in Appendix~\ref{supp:prompt-adjust-process}.

During the self-adaptation process, we first draft an initial template, a class structure with fields, including \texttt{is\_ai\_generated: Bool, raw\_frame\_analysis: str, tool\_name\_analysis: str, explanation: str}. Starting from the first iteration, LVLMs propose new templates based on the aforementioned system prompt.
If a template underperforms, incremental modifications will be applied to key fields in the class-structure of prompts, ensuring adjustments focus on broader analytical aspects, such as facial feature consistency or temporal anomaly detection. 
This iterative refinement improves the adaptability of the model, particularly in challenging real-world datasets.
Our approach not only prevents the model from overfitting to specific words or phrases but also mitigates hallucinations in non-structured prompts. 
The iterative process encourages high-level improvements in classification accuracy rather than focusing on low-level, superficial changes. More details on self-adaptation process, including the whole prompt template and evolution logging are shown in Appdx Sec.~\ref{supp:prompt-adjust-process}

\section{Experiment}
\label{sec:experiment}



\subsection{Experiment Setting}
\paragraph{Model}
We evaluate the \reveal framework using four leading Large Language Vision Models (LVLMs):
\textbf{1) Llava-OV-7B}~\cite{li2024llavaonevisioneasyvisualtask} represents Llava-OneVision-7B, an open-source LVLM known for its strong visual understanding capabilities. This model is selected to test \reveal enhancement on small LVLMs.
\textbf{2) Qwen-VL-Max}~\cite{qwen2023qwenvl} refers to Qwen-VL-Max-0809, a top-performing commercial LVLM from Alibaba~\cite{alibaba}. 
\textbf{3) Gemini-1.5-pro}~\cite{google2024gemini} is one of the most advanced commercial LVLMs from Google. We choose the Gemini-1.5-pro-002 version.
\textbf{4) GPT-4o}~\cite{openai2024gpt4o} is the most advanced LVLM from OpenAI. It offers the structural prompt configuration in our evaluation. We select the GPT-4o-0806 version.

\paragraph{Dataset}
We introduce \vidfor, our video detection benchmark composed of a diverse set of real videos and diffusion-generated videos created by open-source text-to-video generation tools. 
\vidfor consists of selections of videos from PANDA-70M and VidProM datasets. It is enhanced by our in-house combination of real videos sourced from YouTube and generated videos produced by four SOTA text-to-video generation models: Kling~\cite{klingai}, Gen3~\cite{runway3}, SORA~\cite{openaisora}, and OpenSORA~\cite{opensora}. 
\textbf{Kling}, a video generation platform created by KuaiShou~\cite{kuaishou}. With a combination of model architectures, including 3D-VAE, and 3D-spatio-temporal join attention mechanism,  Kling generates high-quality videos (up to two minutes) that conform to physical laws~\cite{klingai}.
\textbf{Gen3}, created by Runway~\cite{runway3}, was trained with a multimodal dataset and released with a set of safeguards. Gen3 produces videos that feature photorealistic human characters with advanced motion and stylistic control.
Developed by OpenAI, \textbf{SORA} is a diffusion-based text-to-video model with a profound understanding of scene complexity and real world objects~\cite{openaisora,videoworldsimulators2024}. \textbf{OpenSORA} is an open-source product of HPC-AI Tech trained on $\sim$30 million data and highlights an innovative video compression network~\cite{opensora}.

\paragraph{Baseline}
\begin{itemize}
    \item \textbf{Direct Prompt} We perform the baseline method for each LVLM by directly prompting itself with the message "if the consecutive frames input is generated by AI or not". To thoroughly evaluate the general performance of these models in video detection, we carefully design three direct prompts, where the placeholder $\{\texttt{prompt p}\}$ can be replaced with prompt \texttt{p1}, \texttt{p2}, and \texttt{p3}. All three baselines have the same meaning, but they are phrased differently. We discover that different paraphrased versions of prompts can lead to different performance of LVLMs.

\end{itemize}
\begin{tcolorbox}[colback=white!5!white,colframe=gray!2!gray,arc=2mm,fontupper=\small, left=2pt, right=2pt]
\begin{adjustbox}{minipage=[c][0.33\linewidth]{1\linewidth}}
Direct Prompt: "These 8 images are consecutive frames of a video. \{\texttt{prompt p}\}. Must return with 1) Yes or No only; 2) if Yes, explain the reason."\\

\texttt{p1.} Do you think this video is generated by AI or not? \\
\texttt{p2.} Tell me if there are synthetic artifacts in the video or not? \\
\texttt{p3.} Do you think this video was created with the help of AI?
\end{adjustbox}
\end{tcolorbox}

\begin{itemize}
\item \textbf{Supervised learning baselines} We compare \reveal with supervised learning detection methods. We extract vision features of all explicit knowledge and train with three methods: XGBoost~\cite{chen2016xgboost}, SVM~\cite{cortes1995support}, and transformer~\cite{vaswani2017attention}. XGBoost is configured with 100 estimators, a maximum depth of 6. The SVM employs a linear kernel that is enabled by probability estimation. The transformer-based method follows a standard encoder structure, consisting of a multi-head self-attention mechanism, a feedforward network, residual connections, layer normalization, and positional encoding. We trained these models using data outside the test set.
\end{itemize}

\paragraph{Implementation Details}
In our experiments, all LVLMs are configured to accept multiple image inputs. 
Videos in \vidfor are all processed to a maximum of 100 consecutive frames. Additionally, for each video, we select the middle 8 frames as input to the model. 
We also test the impact of using the first 8 frames and the last 8 frames on detection results, observing that the results are consistent across these three frame selection methods.
We set the hyperparameters for model generation, such as temperature $T=0$. For online adaptation implementation, we process the adaptation set in batches of 25 examples, using an F1-score threshold of 0.8 to encourage adaptation while maintaining performance standards. We set the adaptation iteration limit to 20. For template re-writing, we provide specific guidance to focus on high-level analysis aspects. For each iteration, we ask the LVLM to propose a new field name in our structured prompt. After each template trial, we record all the rewriting records and the corresponding F1 scores, allowing the LVLM to analyze past results and identify valuable fields for continuous improvement. In Appendix~\ref{supp:prompt_detail}~\ref{supp:prompt-adjust-process}, we show the prompt details for selecting explicit knowledge.


\paragraph{Evaluation Metrics}
In our experiment, we aim for the model to detect artifacts in the additional information that are absent in the original raw video. When LVLMs are presented with a video for detection, using the provided toolkit, they first perform an independent analysis of each explicit knowledge information. Then, we integrate the prediction of each explicit knowledge using an OR operation to ensemble the final result for the video. Video-level accuracy and F1 score are adopted as the evaluation metrics for all experiments.

\begin{table*}[t]
\centering
\scriptsize
\setlength{\tabcolsep}{3pt} 
\renewcommand{\arraystretch}{1.4} 
\begin{tabular}{ll|cccc|ccccc}
\toprule
\multirow{2}{*}{\textbf{LVLM}} & \multirow{2}{*}{\textbf{Method}} & \multicolumn{4}{c|}{\textbf{VidForensic (VidProM)~\cite{wang2024vidprommillionscalerealpromptgallery}}} & \multicolumn{4}{c}{\textbf{VidForensic (Self-collected)}} & \multirow{2}{*}{\textbf{Avg.}} \\ \cline{3-10}
                      &                         & \textbf{Pika~\cite{pika}}   & \textbf{T2vz~\cite{khachatryan2023text2videozerotexttoimagediffusionmodels}}  & \textbf{Vc2~\cite{chen2024videocrafter2overcomingdatalimitations}}  & \textbf{Ms~\cite{wang2023modelscopetexttovideotechnicalreport}}  & \textbf{OpenSORA~\cite{opensora}} & \textbf{Gen3~\cite{runway3}} & \textbf{Kling~\cite{klingai}} & \textbf{SORA~\cite{videoworldsimulators2024}} &                       \\  
\midrule
--- & \textbf{XGBoost}~\cite{chen2016xgboost}    & 56.00/46.55      & 50.50/34.42     & 57.00/48.51    & 52.00/37.98   & 53.00/40.25        & 54.00/42.43    & 55.00/44.53     & 51.11/29.37    & 53.58/40.50                     \\ 
\midrule

                   --- &     \textbf{SVM}~\cite{cortes1995support}  & 63.00/56.65      & 78.00/77.68    & 86.00/85.90    & 85.50/85.43   & 67.50/63.94        & 65.00/60.02    & 68.50/65.43     &52.22/26.48    & 70.72/65.19                     \\ 
\midrule
          
          --- &            \textbf{ViT}~\cite{vaswani2017attention}  & 76.56/76.01      & 88.90/88.51     & 91.52/90.73   & 91.52/90.73   & 81.33/81.29        & 83.34/83.33    & 75.51/74.76     & 61.14/61.08    & 81.23/80.81                    \\ 
                             
\midrule

\multirow{4}{*}{\textbf{Llava-OV-7B}~\cite{li2024llavaonevisioneasyvisualtask}} & Direct prompt 1    & 53.50/14.68      & 61.00/37.10     & 61.00/37.10    & 58.50/30.25   & 52.50/12.11        & 50.00/1.96    & \underline{50.00}/1.96     & 54.44/16.33    & 55.12/18.94                     \\ 

                             & Direct prompt 2        & 50.50/1.98      & 51.00/3.92    & 51.50/5.83    & 53.50/13.08   & 52.00/7.69        & 50.00/0.00    & 50.00/0.00     & 50.00/0.00    & 51.06/4.06                     \\ 
                             
                             & Direct prompt 3         & \textbf{54.50}/\textbf{18.02}      & \underline{62.00}/\underline{39.68}     & \underline{65.00}/\underline{46.97}    & \underline{62.00}/\underline{39.68}   & \underline{54.00}/\underline{16.36}        & \underline{51.00}/\underline{5.77}    & 50.00/\underline{1.96}     & \underline{55.56}/\underline{20.00}    & \underline{56.76}/\underline{23.56}                     \\ 
                             
\cline{2-11}
                             & \textbf{\reveal (ours)}            & \textbf{54.50}/\textbf{18.02}     & \textbf{70.00}/\textbf{57.75}     & \textbf{69.00}/\textbf{55.71}    & \textbf{68.00}/\textbf{53.62}   & \textbf{58.00}/\textbf{28.81}        & \textbf{51.50}/\textbf{7.62}    & \textbf{50.50}/\textbf{3.88}     & \textbf{55.56}/\textbf{20.00}    & \textbf{59.63}/\textbf{32.69}                     \\
\midrule
                             
\multirow{4}{*}{\textbf{Qwen-VL-Max}~\cite{qwen2023qwenvl}} & Direct prompt 1     & 72.50/63.09      & 75.00/67.53     & 82.00/78.57    & 76.00/69.23   & 67.50/53.24        & 62.00/40.62    & \underline{54.50}/19.47     & 58.89/39.34    & 68.55/51.24                     \\ 

                             & Direct prompt 2       & 60.50/38.76      & 75.00/68.35    & 71.50/62.25    & 72.50/64.05   & 60.50/38.76        & 52.00/14.29    & 50.00/7.41     & 56.67/26.42    & 62.33/39.56                     \\ 
                             
                             & Direct prompt 3        & \underline{74.00}/\underline{67.90}      & \underline{79.00}/\underline{75.58}     & \underline{84.50}/\underline{83.06}    & \textbf{79.50}/\underline{76.30}   & \underline{69.50}/\underline{60.13}        & \underline{65.50}/\underline{52.41}    & 54.00/\underline{24.59}     & \underline{61.11}/\underline{47.76}    & \underline{70.89}/\underline{60.97}                     \\ 
                             
\cline{2-11}
                             & \textbf{\reveal (ours)}         & \textbf{87.00}/\textbf{88.39}     & \textbf{81.50}/\textbf{82.63}     & \textbf{86.00}/\textbf{87.39}    & \underline{77.00}/\textbf{77.45}   & \textbf{79.00}/\textbf{79.81}        & \textbf{82.50}/\textbf{83.72}    & \textbf{60.00}/\textbf{52.94}     & \textbf{67.78}/\textbf{71.84}    & \textbf{77.60}/\textbf{76.08}                     \\

\midrule
\multirow{4}{*}{\textbf{Gemini-1.5-pro}~\cite{google2024gemini}} & Direct prompt 1        & 68.33/54.32      & 71.00/59.72     & 67.00/51.47    & 75.00/67.11   & 68.50/54.68        & 64.00/44.62    & 58.00/28.81    & 58.89/41.27    & \underline{66.34}/49.83                     \\ 

                             & Direct prompt 2        & \underline{73.50}/\underline{66.24}      & \underline{81.00}/\underline{77.91}     & \underline{76.00}/\underline{70.37}    & \underline{85.00}/\underline{83.33}  & \underline{71.50}/\underline{62.75}       & \underline{71.50}/\underline{62.75}  & \underline{59.50}/\underline{37.21}     & \underline{71.11}/\underline{64.86}    & 72.51/\underline{58.28}                     \\ 
                             
                             & Direct prompt 3       & 64.50/45.80      & 77.00/70.51     & 71.00/59.72    & 76.50/69.68   & 64.50/45.80        & 62.00/39.68    & 52.50/11.21     & 61.11/42.62    & 66.08/51.28                     \\ 
                             
\cline{2-11}
                             & \textbf{\reveal (ours)}           & \textbf{92.00}/\textbf{91.73}     & \textbf{96.33}/\textbf{96.38}    & \textbf{95.83}/\textbf{95.87}    & \textbf{97.50}/\textbf{97.56}   & \textbf{92.17}/\textbf{91.93}        & \textbf{88.50}/\textbf{87.67}    & \textbf{74.83}/\textbf{68.46}     & \textbf{76.67}/\textbf{78.36}    & \textbf{89.23}/\textbf{88.43}                     \\
\midrule

\multirow{5}{*}{\textbf{GPT-4o}~\cite{openai2024gpt4o}}      & Direct prompt 1      & 89.00/89.22     & 90.00/90.29     & \textbf{92.50}/92.89    & 85.00/84.69   & 82.50/81.68        & 86.00/85.86    & 66.50/57.86     & \underline{68.89}/64.10    & 82.55/80.82                     \\  

                             & Direct prompt 2     & 72.00/77.95      & 70.00/76.00     & 71.00/76.98    & 66.50/72.43   & 68.00/73.98        & 68.00/73.98    & 64.50/70.29   & 65.56/\underline{70.84}   & 68.20/74.06                     \\ 
                             
                             & Direct prompt 3      & 89.50/88.66      & 90.50/90.73     & 92.00/92.31    & 86.00/85.71  & 82.00/80.85        & 85.00/84.54    & 69.00/61.73     & 63.33/50.75   & 82.17/79.41                     \\  
                             
\cline{2-11}
                             
                             & \textbf{\reveal (ours)}   & 
                             \underline{91.50}/\underline{92.17}      & \textbf{92.00}/\textbf{92.52}     & \textbf{92.50}/\textbf{93.02}   & \textbf{90.50}/\textbf{91.24}   & \textbf{86.50}/\textbf{86.79}        & \textbf{91.00}/\textbf{91.59}    &  \textbf{77.00}/\textbf{76.77}& \textbf{70.93}/\textbf{72.11}    & \textbf{86.49}/\textbf{87.03}  
                             \\         
\bottomrule
\end{tabular}

\caption{Performance comparison of \reveal with three supervised learning-based detection methods and three direct prompting methods across eight datasets. For each dataset except SORA, we combine the real dataset from Panda-70M \& AI-generated dataset together. For SORA, we combine them with 45 YouTube videos that collected by ourselves. We use four representative LVLMs to serve as the detector in our framework, including Llava-OV-7B~\cite{li2024llavaonevisioneasyvisualtask}, Qwen-VL-Max~\cite{qwen2023qwenvl}, Gemini-1.5-pro~\cite{google2024gemini}, and GPT-4o~\cite{openai2024gpt4o}. The results are presented as Accuracy / F1-score in each cell. Numbers in bold show the top-1 best results, and numbers with underlined show the top-2 best results.}
\label{tab:non-str-results}
\end{table*}

\subsection{Experimental Results}
In Table~\ref{tab:non-str-results}, the experiment is conducted with four LVLM models.
The result demonstrates that our \reveal framework could consistently surpass its baseline setting with zero-shot prompt across all eight datasets. For Qwen-VL-Max~\cite{qwen2023qwenvl} and Gemini-1.5-pro~\cite{google2024gemini}, compared to the best-performing baselines, \reveal outperforms them on average F1 score by \textbf{15.1}\% and \textbf{30.2\%} gain. 
For Llava-OV-7B, the average F1 score slightly improves by 9.13\% points across all eight datasets, compared to baselines.  
We believe this outcome is because the model capacity of Llava-OV-7B is too small and has limited understanding of explicit knowledge. 
We show the result of GPT-4o~\cite{openai2024gpt4o} with structured prompt. Although GPT-4o demonstrates impressive multimodal performance and is the highest-performing baseline (achieving an average F1 of 80.8\%) among all models, \reveal still outperforms it with an average improvement of \textbf{6.2\%} across all datasets and a stable average improvement of \textbf{9.4\%} on the high-quality VidForensic subsets.
In Appendix~\ref{supp:more_results}, we show additional results on video-specific settings.



\begin{table}[t]
\centering
\setlength{\tabcolsep}{1pt} 
\scriptsize
\begin{tabular}{@{}l|ccccccccc@{}}
\toprule
\textbf{Model} & \textbf{Land.} & \textbf{Depth} & \textbf{Enhan.} & \textbf{Edge} & \textbf{Sharp.} & \textbf{Denoise} & \textbf{OPflow} & \textbf{Sat.} & \textbf{SAM}\\ \midrule
Llava-OV-7B~\cite{li2024llavaonevisioneasyvisualtask} &  &  & \checkmark &  & & & \checkmark & \checkmark   \\
Qwen-VL-Max~\cite{qwen2023qwenvl}  & \checkmark & \checkmark & & & \checkmark & & \checkmark &    \\
Gemini-1.5-pro~\cite{google2024gemini} & \checkmark & \checkmark & \checkmark & \checkmark &  & \checkmark & \checkmark &  \\ 
GPT-4o~\cite{openai2024gpt4o} & \checkmark & & & \checkmark &  & &  & \checkmark  \\ 

\bottomrule
\end{tabular}
\caption{Summarization of model-specific explicit knowledge tools. The self-evolving selection strategy in \reveal helps different LVLMs to customize their own useful tools for video detection.}
\label{tab:ek_tool_set}
\end{table}

\section{Ablation Studies}

\label{sec:abalation}

\subsection{Analysis on Deepfake video detection}

Recent work~\cite{jia2024chatgptdetectDeepfakesstudy} shows that LVLMs can be effectively applied to Deepfake detection tasks.
To investigate this, we adopt \reveal to Gemini-1.5-Pro~\cite{google2024gemini} and GPT-4o on Celeb-DF-v1~\cite{Celeb_DF_cvpr20}, a Deepfake dataset. 
In Table~\ref{tab:abalation_study_Deepfake}, we compare \reveal with three deep learning-based baselines~\cite{9878441, guo2023controllableguidespacegeneralizableface, zhao2021multiattentionalDeepfakedetection} trained on FaceForensics++ ~\cite{roessler2019faceforensicspp} (FF++). 
Additionally, prior work~\cite{jia2024chatgptdetectDeepfakesstudy} shows that decomposed facial features can potentially improve Deepfake detection. Therefore, we utilize open-source tool, \emph{Language Segment-Anything}\footnote{Language Segment-Anything: \url{https://github.com/luca-medeiros/lang-segment-anything}.} to segment facial features (Face-Seg), treating it as an additional explicit knowledge for \reveal. In Table~\ref{tab:abalation_study_Deepfake},
we observe that \reveal(GPT-4o) demonstrates comparable performance to baseline methods by achieving 75.0\% video-level detection accuracy. Compared to baseline prompting approaches, \reveal improves Gemini-1.5-Pro~\cite{google2024gemini} by 6.0\% in accuracy and 19.85\% in F1-score. It also increases GPT-4o accuracy by 10.05\% and F1-score by 6.67\%. This study demonstrates the capabilities of \reveal in Deepfake detection.

\begin{figure*}[t]
    \centering
    \includegraphics[width=0.95\textwidth]{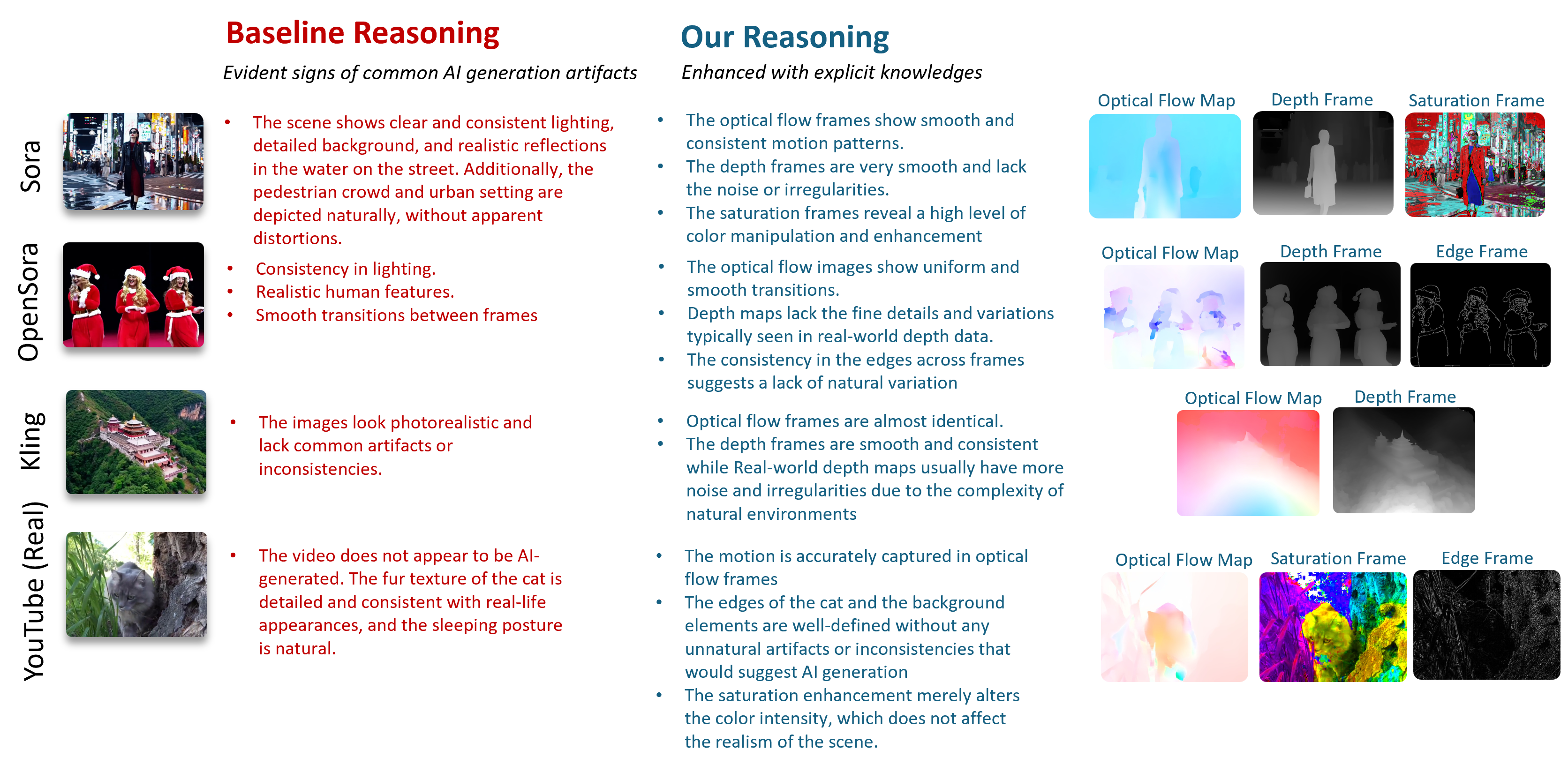}
    \vspace{-2mm}
    \caption{Reasoning visualization for \textbf{\reveal}.  We visualize three AI-generated samples and one real-world sample. The baseline reasoning illustrates the output reasoning when we directly query the LVLM with raw video and direct prompt. Our reasoning represents the output reasoning when we query LVLM with multiple explicit knowledges and structured prompt.}
    \label{fig:reasoning}
\end{figure*}

\begin{table}[t]
\centering
\setlength{\tabcolsep}{2pt} 
\tiny
\resizebox{0.85\columnwidth}{!}{%
\begin{tabular}{@{}l|c|cc@{}}
\hline
\multirow{2}{*}{\textbf{Method}}        & \multirow{2}{*}{\textbf{Trainset}} & \multicolumn{2}{c}{\textbf{Celeb-DF-v1}} \\
\cline{3-4}
                   &  & \textbf{Acc.} & \textbf{F1} \\ \hline
Guo et al.~\cite{guo2023controllableguidespacegeneralizableface}            &   FF++ ~\cite{roessler2019faceforensicspp}    &   73.19    &      --       \\
RECCE~\cite{9878441}               &   FF++ ~\cite{roessler2019faceforensicspp}  &   71.81    &     --       \\
MAT~\cite{zhao2021multiattentionalDeepfakedetection}            &   FF++ ~\cite{roessler2019faceforensicspp}    &   71.81    &    --      \\
\hline
Baseline (Gemini-1.5-pro)              &  --  &  44.00    &  17.65           \\
Baseline (GPT-4o)             &  --  &  64.95    &   74.24          \\
\hline
\textbf{\textit{\reveal (Gemini-1.5-pro) w/ Face-Seg}}          &   --   &   50.00   &    37.50       \\
\textbf{\textit{\reveal (GPT-4o) w/ Face-Seg}}             &   --   &   \textbf{75.00}   &    \textbf{80.91}       \\
\hline
\end{tabular}%
}

\caption{Performance comparison of existing Deepfake detection baselines, the baseline prompts, and \reveal on Celeb-DF-v1. Video-level accuracy (Acc.) and F1-score (F1) are used as evaluation metrics where available. The reported performance of RECCE and MAT are referenced from~\cite{Wang_2024}.}
\label{tab:abalation_study_Deepfake}
\vspace{-2mm}
\end{table}

\subsection{Reasoning Visualization}
To demonstrate the reasoning capabilities of \reveal, we show how the model explains if a video is real or not. We observe that it is meaningful to determine which characteristics the model considers to be important for recognizing the positive samples and whether these descriptions are reasonable. In Fig~\ref{fig:reasoning}, we compare the reasoning output from LVLM via direct prompt and our method \reveal. We show four examples, including three AI-generated videos from Sora, OpenSora, and Kling, and one real video from YouTube. Unlike the direct prompts, which provide baseline reasoning and exhibit general AI-generated artifacts as indicators, \reveal offers detailed rationale based on the corresponding explicit knowledge. We observe that those subtle unnatural physics that might only happen in AI-generated videos are well-captured by \reveal, such as frames of depth map being overly smooth or optical flow map being almost identical in every frame.

\section{Discussion}
\paragraph{Dependency on LVLMs} Based on our thorough investigation, we confirm that the reasoning capabilities of current state-of-the-art (SOTA) LVLMs are inadequate. Our proposed solution significantly enhances LVLMs by enabling the selection of explicit knowledge and the detection of artifacts. As LVLMs continue to evolve and integrate greater reasoning abilities, it is imperative to explore how to guide these models to specific reasoning and employ chain-of-thought reasoning strategies into self-evolving process.
\paragraph{Computational Cost for Detection} To further investigate 
the extent of the computational cost for using LVLMs for detection, in Table~\ref{tab:cost}, we summarize the cost for three different commercial LVLMs, including GPT-4o, Gemini 1.5 pro, and QWen-VL-Max. The video cost is calculated by \# of text token $\times$ \$ of text token + \# of frames $\times$ \$ of input frame. Here, the \# of frame is set as eight.

\begin{table}[h!]
\vspace{-2mm}
\centering
\scriptsize
\renewcommand{\arraystretch}{0.5} 
\begin{tabular}{@{}lccc@{}}
\toprule
Name             & Input Text Token & 1080p Input Frame & Input Video\\ \midrule
GPT4o-0806       & \$2.50/1M                   & \$2.763/1K & \$22.215/1K                                          \\
Gemini-1.5-pro-002 & \$1.25/1M                  & \$0.323/1K  & \$2.639/1K                                       \\
Qwen-VL-MAX-0809 & \$2.78/1M         & \$3.451/1K   & \$27.728/1K                                     \\ \bottomrule
\end{tabular}
\caption{Computation cost of \reveal}
\vspace{-3mm}
\label{tab:cost}
\end{table}

\section{Conclusion}
\label{sec:conclusion}

\reveal is a novel agentic framework that leverages LVLMs' strong reasoning ability to detect diffusion-generated video. 
As opposed to existing methods that require supervised training detectors with explicit knowledge (EK), \reveal is training-free and can generalize to videos produced by different sources of video generation tools.
With our proposed EK selection method based on a tool-preference metric, \reveal can effectively extract useful EK for LVLMs to perform detection. We further propose a self-adaptation method for structured prompts based on a rewriting template mechanism. Our proposed process prevents LVLMs from overfitting with a specific template. 
The evaluation demonstrates that \reveal improves F1 scores by 9.1\% to 30.2\% over the top baseline on a high-quality video dataset across four leading LVLMs. Our work offers fresh perspectives on video detection by employing an agentic LVLM framework with emerging techniques.



{
    \small
    \bibliographystyle{ieeenat_fullname}
    \bibliography{main}

@String(CVPR= {IEEE Conf. Comput. Vis. Pattern Recog.})

@String(ICCV= {Int. Conf. Comput. Vis.})

@String(CVPR  = {CVPR})

@String(ICCV  = {ICCV})

@article{cortes1995support,
  title={Support-vector networks},
  author={Cortes, Corinna and Vapnik, Vladimir N.},
  journal={Machine learning},
  volume={20},
  number={3},
  pages={273--297},
  year={1995},
  publisher={Springer}
}

@inproceedings{chen2016xgboost,
  title={XGBoost: A Scalable Tree Boosting System},
  author={Chen, Tianqi and Guestrin, Carlos},
  booktitle={Proceedings of the 22nd ACM SIGKDD International Conference on Knowledge Discovery and Data Mining},
  pages={785--794},
  year={2016},
  organization={ACM}
}

@inproceedings{vaswani2017attention,
  title={Attention Is All You Need},
  author={Vaswani, Ashish and Shazeer, Noam and Parmar, Niki and Uszkoreit, Jakob and Jones, Llion and Gomez, Aidan N and Kaiser, Lukasz and Polosukhin, Illia},
  booktitle={Advances in Neural Information Processing Systems},
  pages={5998--6008},
  year={2017}
}

@misc{chen2024panda70mcaptioning70mvideos,
      title={Panda-70M: Captioning 70M Videos with Multiple Cross-Modality Teachers}, 
      author={Tsai-Shien Chen and Aliaksandr Siarohin and Willi Menapace and Ekaterina Deyneka and Hsiang-wei Chao and Byung Eun Jeon and Yuwei Fang and Hsin-Ying Lee and Jian Ren and Ming-Hsuan Yang and Sergey Tulyakov},
      year={2024},
      eprint={2402.19479},
      archivePrefix={arXiv},
      primaryClass={cs.CV},
      url={https://arxiv.org/abs/2402.19479}, 
}

@misc{liu2024turnsimrealrobust,
      title={Turns Out I'm Not Real: Towards Robust Detection of AI-Generated Videos}, 
      author={Qingyuan Liu and Pengyuan Shi and Yun-Yun Tsai and Chengzhi Mao and Junfeng Yang},
      year={2024},
      eprint={2406.09601},
      archivePrefix={arXiv},
      primaryClass={cs.CV},
      url={https://arxiv.org/abs/2406.09601}, 
}

@misc{wang2024vidprommillionscalerealpromptgallery,
      title={VidProM: A Million-scale Real Prompt-Gallery Dataset for Text-to-Video Diffusion Models}, 
      author={Wenhao Wang and Yi Yang},
      year={2024},
      eprint={2403.06098},
      archivePrefix={arXiv},
      primaryClass={cs.CV},
      url={https://arxiv.org/abs/2403.06098}, 
}

@misc{chen2024videocrafter2overcomingdatalimitations,
      title={VideoCrafter2: Overcoming Data Limitations for High-Quality Video Diffusion Models}, 
      author={Haoxin Chen and Yong Zhang and Xiaodong Cun and Menghan Xia and Xintao Wang and Chao Weng and Ying Shan},
      year={2024},
      eprint={2401.09047},
      archivePrefix={arXiv},
      primaryClass={cs.CV},
      url={https://arxiv.org/abs/2401.09047}, 
}

@software{opensora,
  author = {Zangwei Zheng and Xiangyu Peng and Tianji Yang and Chenhui Shen and Shenggui Li and Hongxin Liu and Yukun Zhou and Tianyi Li and Yang You},
  title = {Open-Sora: Democratizing Efficient Video Production for All},
  month = {March},
  year = {2024},
  url = {https://github.com/hpcaitech/Open-Sora}
}

@misc{wang2023modelscopetexttovideotechnicalreport,
      title={ModelScope Text-to-Video Technical Report}, 
      author={Jiuniu Wang and Hangjie Yuan and Dayou Chen and Yingya Zhang and Xiang Wang and Shiwei Zhang},
      year={2023},
      eprint={2308.06571},
      archivePrefix={arXiv},
      primaryClass={cs.CV},
      url={https://arxiv.org/abs/2308.06571}, 
}

@misc{khachatryan2023text2videozerotexttoimagediffusionmodels,
      title={Text2Video-Zero: Text-to-Image Diffusion Models are Zero-Shot Video Generators}, 
      author={Levon Khachatryan and Andranik Movsisyan and Vahram Tadevosyan and Roberto Henschel and Zhangyang Wang and Shant Navasardyan and Humphrey Shi},
      year={2023},
      eprint={2303.13439},
      archivePrefix={arXiv},
      primaryClass={cs.CV},
      url={https://arxiv.org/abs/2303.13439}, 
}

@article{videoworldsimulators2024,
  title={Video generation models as world simulators},
  author={Tim Brooks and Bill Peebles and Connor Holmes and Will DePue and Yufei Guo and Li Jing and David Schnurr and Joe Taylor and Troy Luhman and Eric Luhman and Clarence Ng and Ricky Wang and Aditya Ramesh},
  year={2024},
  url={https://openai.com/research/video-generation-models-as-world-simulators},
}

@misc{runway3,
title={Runway ML. Gen2. (2024) [Text-to-video model]. https://research.runwayml.com/}, url={https://research.runwayml.com/}}

@misc{pika,
title={Pika Labs. (2024) [Text-to-video model]. https://www.pika.art/} ,url={https://www.pika.art/}
}

@misc{klingai,
  title        = {Kling AI},
  year         = {2024},
  note         = {[Text-to-video model]}, 
  url          = {https://www.klingai.com/}
}

@misc{openaisora,
  author       = {OpenAI},
  title        = {Sora},
  year         = {2024},
  url         = {https://openai.com/index/sora/}
}

@misc{alibaba,
  author       = {Alibaba},
  title        = {Alibaba},
  year         = {2024},
  url         = {alibaba.com}
}

@misc{kuaishou,
  author       = {kuaishou},
  title        = {kuaishou},
  year         = {2024},
  url         = {kuaishou.com}
}

@article{zhang2023show,
  title={Show-1: Marrying Pixel and Latent Diffusion Models for Text-to-Video Generation},
  author={Zhang, David Junhao and Wu, Jay Zhangjie and Liu, Jia-Wei and Zhao, Rui and Ran, Lingmin and Gu, Yuchao and Gao, Difei and Shou, Mike Zheng},
  journal={arXiv preprint arXiv:2309.15818},
  year={2023}
}

@misc{blattmann2023stable,
      title={Stable Video Diffusion: Scaling Latent Video Diffusion Models to Large Datasets}, 
      author={Andreas Blattmann and Tim Dockhorn and Sumith Kulal and Daniel Mendelevitch and Maciej Kilian and Dominik Lorenz and Yam Levi and Zion English and Vikram Voleti and Adam Letts and Varun Jampani and Robin Rombach},
      year={2023},
      eprint={2311.15127},
      archivePrefix={arXiv},
      primaryClass={cs.CV}
}

@article{deepfake_pavel,
  author       = {Pavel Korshunov and
                  S{\'{e}}bastien Marcel},
  title        = {DeepFakes: a New Threat to Face Recognition? Assessment and Detection},
  journal      = {CoRR},
  volume       = {abs/1812.08685},
  year         = {2018},
  url          = {http://arxiv.org/abs/1812.08685},
  eprinttype    = {arXiv},
  eprint       = {1812.08685},
  bibsource    = {dblp computer science bibliography, https://dblp.org}
}

@INPROCEEDINGS{8639163_deepfake,
  author={Güera, David and Delp, Edward J.},
  booktitle={2018 15th IEEE International Conference on Advanced Video and Signal Based Surveillance (AVSS)}, 
  title={Deepfake Video Detection Using Recurrent Neural Networks}, 
  year={2018},
  volume={},
  number={},
  pages={1-6},
  doi={10.1109/AVSS.2018.8639163}}

@misc{radford2021learningtransferablevisualmodels,
      title={Learning Transferable Visual Models From Natural Language Supervision}, 
      author={Alec Radford and Jong Wook Kim and Chris Hallacy and Aditya Ramesh and Gabriel Goh and Sandhini Agarwal and Girish Sastry and Amanda Askell and Pamela Mishkin and Jack Clark and Gretchen Krueger and Ilya Sutskever},
      year={2021},
      eprint={2103.00020},
      archivePrefix={arXiv},
      primaryClass={cs.CV},
      url={https://arxiv.org/abs/2103.00020}, 
}

@misc{jia2021scalingvisualvisionlanguagerepresentation,
      title={Scaling Up Visual and Vision-Language Representation Learning With Noisy Text Supervision}, 
      author={Chao Jia and Yinfei Yang and Ye Xia and Yi-Ting Chen and Zarana Parekh and Hieu Pham and Quoc V. Le and Yunhsuan Sung and Zhen Li and Tom Duerig},
      year={2021},
      eprint={2102.05918},
      archivePrefix={arXiv},
      primaryClass={cs.CV},
      url={https://arxiv.org/abs/2102.05918}, 
}

@misc{alayrac2022flamingovisuallanguagemodel,
      title={Flamingo: a Visual Language Model for Few-Shot Learning}, 
      author={Jean-Baptiste Alayrac and Jeff Donahue and Pauline Luc and Antoine Miech and Iain Barr and Yana Hasson and Karel Lenc and Arthur Mensch and Katie Millican and Malcolm Reynolds and Roman Ring and Eliza Rutherford and Serkan Cabi and Tengda Han and Zhitao Gong and Sina Samangooei and Marianne Monteiro and Jacob Menick and Sebastian Borgeaud and Andrew Brock and Aida Nematzadeh and Sahand Sharifzadeh and Mikolaj Binkowski and Ricardo Barreira and Oriol Vinyals and Andrew Zisserman and Karen Simonyan},
      year={2022},
      eprint={2204.14198},
      archivePrefix={arXiv},
      primaryClass={cs.CV},
      url={https://arxiv.org/abs/2204.14198}, 
}

@misc{zellers2021merlotmultimodalneuralscript,
      title={MERLOT: Multimodal Neural Script Knowledge Models}, 
      author={Rowan Zellers and Ximing Lu and Jack Hessel and Youngjae Yu and Jae Sung Park and Jize Cao and Ali Farhadi and Yejin Choi},
      year={2021},
      eprint={2106.02636},
      archivePrefix={arXiv},
      primaryClass={cs.CV},
      url={https://arxiv.org/abs/2106.02636}, 
}

@misc{li2023blip2bootstrappinglanguageimagepretraining,
      title={BLIP-2: Bootstrapping Language-Image Pre-training with Frozen Image Encoders and Large Language Models}, 
      author={Junnan Li and Dongxu Li and Silvio Savarese and Steven Hoi},
      year={2023},
      eprint={2301.12597},
      archivePrefix={arXiv},
      primaryClass={cs.CV},
      url={https://arxiv.org/abs/2301.12597}, 
}

@article{wang2023dire,
  title={DIRE for Diffusion-Generated Image Detection},
  author={Wang, Zhendong and Bao, Jianmin and Zhou, Wengang and Wang, Weilun and Hu, Hezhen and Chen, Hong and Li, Houqiang},
  journal={arXiv preprint arXiv:2303.09295},
  year={2023}
}

@misc{taylorswift,
title={Look What You Made Me Do: Why Deepfake Taylor Swift Matters},
year={2024}, url ={     https://www.forbes.com/sites/falonfatemi/2024/02/01/look-what-you-made-me-do-why-deepfake-taylor-swift-matters/}}

@misc{zhong2022proqastructuralpromptbasedpretraining,
      title={ProQA: Structural Prompt-based Pre-training for Unified Question Answering}, 
      author={Wanjun Zhong and Yifan Gao and Ning Ding and Yujia Qin and Zhiyuan Liu and Ming Zhou and Jiahai Wang and Jian Yin and Nan Duan},
      year={2022},
      eprint={2205.04040},
      archivePrefix={arXiv},
      primaryClass={cs.CL},
      url={https://arxiv.org/abs/2205.04040}, 
}

@misc{jia2024chatgptdetectdeepfakesstudy,
      title={Can ChatGPT Detect DeepFakes? A Study of Using Multimodal Large Language Models for Media Forensics}, 
      author={Shan Jia and Reilin Lyu and Kangran Zhao and Yize Chen and Zhiyuan Yan and Yan Ju and Chuanbo Hu and Xin Li and Baoyuan Wu and Siwei Lyu},
      year={2024},
      eprint={2403.14077},
      archivePrefix={arXiv},
      primaryClass={cs.AI},
      url={https://arxiv.org/abs/2403.14077}, 
}

@misc{chang2024mattersaigeneratedvideos,
      title={What Matters in Detecting AI-Generated Videos like Sora?}, 
      author={Chirui Chang and Zhengzhe Liu and Xiaoyang Lyu and Xiaojuan Qi},
      year={2024},
      eprint={2406.19568},
      archivePrefix={arXiv},
      primaryClass={cs.CV},
      url={https://arxiv.org/abs/2406.19568}, 
}

@misc{guo2023controllableguidespacegeneralizableface,
      title={Controllable Guide-Space for Generalizable Face Forgery Detection}, 
      author={Ying Guo and Cheng Zhen and Pengfei Yan},
      year={2023},
      eprint={2307.14039},
      archivePrefix={arXiv},
      primaryClass={cs.CV},
      url={https://arxiv.org/abs/2307.14039}, 
}

@misc{chang2024mattersdetectingaigeneratedvideos,
      title={What Matters in Detecting AI-Generated Videos like Sora?}, 
      author={Chirui Chang and Zhengzhe Liu and Xiaoyang Lyu and Xiaojuan Qi},
      year={2024},
      eprint={2406.19568},
      archivePrefix={arXiv},
      primaryClass={cs.CV},
      url={https://arxiv.org/abs/2406.19568}, 
}

@INPROCEEDINGS{9878441,
  author={Cao, Junyi and Ma, Chao and Yao, Taiping and Chen, Shen and Ding, Shouhong and Yang, Xiaokang},
  booktitle={2022 IEEE/CVF Conference on Computer Vision and Pattern Recognition (CVPR)}, 
  title={End-to-End Reconstruction-Classification Learning for Face Forgery Detection}, 
  year={2022},
  volume={},
  number={},
  pages={4103-4112},
  doi={10.1109/CVPR52688.2022.00408}}

@misc{zhao2021multiattentionaldeepfakedetection,
      title={Multi-attentional Deepfake Detection}, 
      author={Hanqing Zhao and Wenbo Zhou and Dongdong Chen and Tianyi Wei and Weiming Zhang and Nenghai Yu},
      year={2021},
      eprint={2103.02406},
      archivePrefix={arXiv},
      primaryClass={cs.CV},
      url={https://arxiv.org/abs/2103.02406}, 
}

@article{Wang_2024,
   title={Deepfake Detection: A Comprehensive Survey from the Reliability Perspective},
   volume={57},
   ISSN={1557-7341},
   url={http://dx.doi.org/10.1145/3699710},
   DOI={10.1145/3699710},
   number={3},
   journal={ACM Computing Surveys},
   publisher={Association for Computing Machinery (ACM)},
   author={Wang, Tianyi and Liao, Xin and Chow, Kam Pui and Lin, Xiaodong and Wang, Yinglong},
   year={2024},
   month=nov, pages={1–35} }

@misc{chang2024matters,
      title={What Matters in Detecting AI-Generated Videos like Sora?}, 
      author={Chirui Chang and Zhengzhe Liu and Xiaoyang Lyu and Xiaojuan Qi},
      year={2024},
      eprint={2406.19568},
      archivePrefix={arXiv},
      primaryClass={cs.CV},
      url={https://arxiv.org/abs/2406.19568}, 
}

@inproceedings{wang2019cnngenerated,
  title={CNN-generated images are surprisingly easy to spot...for now},
  author={Wang, Sheng-Yu and Wang, Oliver and Zhang, Richard and Owens, Andrew and Efros, Alexei A},
  booktitle={CVPR},
  year={2020}
}

@article{patchforeccv20,
  author       = {Lucy Chai and
                  David Bau and
                  Ser{-}Nam Lim and
                  Phillip Isola},
  title        = {What makes fake images detectable? Understanding properties that generalize},
  journal      = {CoRR},
  volume       = {abs/2008.10588},
  year         = {2020},
  url          = {https://arxiv.org/abs/2008.10588},
  eprinttype    = {arXiv},
  eprint       = {2008.10588},
  bibsource    = {dblp computer science bibliography, https://dblp.org}
}

@inproceedings{Celeb_DF_cvpr20,
   author = {Yuezun Li and Xin Yang and Pu Sun and Honggang Qi and Siwei Lyu},
   title = {Celeb-DF: A Large-scale Challenging Dataset for DeepFake Forensics},
   booktitle= {IEEE Conference on Computer Vision and Patten Recognition (CVPR)},
   year = {2020}
}

@inproceedings{roessler2019faceforensicspp,
	author = {Andreas R\"ossler and Davide Cozzolino and Luisa Verdoliva and Christian Riess and Justus Thies and Matthias Nie{\ss}ner},
	title = {Face{F}orensics++: Learning to Detect Manipulated Facial Images},
	booktitle= {International Conference on Computer Vision (ICCV)},
	year = {2019}
}

@article{openaistructured,
title={Introducing Structured Outputs in the API},
author={OpenAI},
url={https://openai.com/index/introducing-structured-outputs-in-the-api/},
year={2024}
}

@misc{qwen2023qwenvl,
  title        = {Qwen-VL: A Versatile Vision-Language Model for Understanding, Localization, Text Recognition, and Beyond},
  author       = {{Qwen Team}},
  year         = {2023},
  eprint       = {2308.12966},
  archivePrefix= {arXiv},
  primaryClass = {cs.CV}
}

@misc{google2024gemini,
  title        = {Gemini: A Family of Highly Capable Multimodal Models},
  author       = {Google},
  year         = {2024},
  howpublished = {\url{https://cloud.google.com/vertex-ai/generative-ai/docs/model-reference/inference}}
}

@misc{openai2024gpt4o,
  title        = {GPT-4o: Multilingual, Multimodal Generative Pre-trained Transformer},
  author       = {OpenAI},
  year         = {2024},
  howpublished = {\url{https://openai.com}}
}

@misc{blattmann2023stablevideodiffusionscaling,
      title={Stable Video Diffusion: Scaling Latent Video Diffusion Models to Large Datasets}, 
      author={Andreas Blattmann and Tim Dockhorn and Sumith Kulal and Daniel Mendelevitch and Maciej Kilian and Dominik Lorenz and Yam Levi and Zion English and Vikram Voleti and Adam Letts and Varun Jampani and Robin Rombach},
      year={2023},
      eprint={2311.15127},
      archivePrefix={arXiv},
      primaryClass={cs.CV},
      url={https://arxiv.org/abs/2311.15127}, 
}

@misc{chen2024demambaaigeneratedvideodetection,
      title={DeMamba: AI-Generated Video Detection on Million-Scale GenVideo Benchmark}, 
      author={Haoxing Chen and Yan Hong and Zizheng Huang and Zhuoer Xu and Zhangxuan Gu and Yaohui Li and Jun Lan and Huijia Zhu and Jianfu Zhang and Weiqiang Wang and Huaxiong Li},
      year={2024},
      eprint={2405.19707},
      archivePrefix={arXiv},
      primaryClass={cs.CV},
      url={https://arxiv.org/abs/2405.19707}, 
}

@misc{bai2024aigeneratedvideodetectionspatiotemporal,
      title={AI-Generated Video Detection via Spatio-Temporal Anomaly Learning}, 
      author={Jianfa Bai and Man Lin and Gang Cao},
      year={2024},
      eprint={2403.16638},
      archivePrefix={arXiv},
      primaryClass={cs.CV},
      url={https://arxiv.org/abs/2403.16638}, 
}

@misc{ji2024distinguishfakevideosunleashing,
      title={Distinguish Any Fake Videos: Unleashing the Power of Large-scale Data and Motion Features}, 
      author={Lichuan Ji and Yingqi Lin and Zhenhua Huang and Yan Han and Xiaogang Xu and Jiafei Wu and Chong Wang and Zhe Liu},
      year={2024},
      eprint={2405.15343},
      archivePrefix={arXiv},
      primaryClass={cs.CV},
      url={https://arxiv.org/abs/2405.15343}, 
}

@misc{li2024llavaonevisioneasyvisualtask,
      title={LLaVA-OneVision: Easy Visual Task Transfer}, 
      author={Bo Li and Yuanhan Zhang and Dong Guo and Renrui Zhang and Feng Li and Hao Zhang and Kaichen Zhang and Peiyuan Zhang and Yanwei Li and Ziwei Liu and Chunyuan Li},
      year={2024},
      eprint={2408.03326},
      archivePrefix={arXiv},
      primaryClass={cs.CV},
      url={https://arxiv.org/abs/2408.03326}, 
}

@unknown{distinguisgFakeDataset,
author = {Ji, Lichuan and Lin, Yingqi and Huang, Zhenhua and Han, Yan and Wu, Jiafei and Wang, Chong and Liu, Zhe},
year = {2024},
month = {05},
pages = {},
title = {Distinguish Any Fake Videos: Unleashing the Power of Large-scale Data and Motion Features},
doi = {10.48550/arXiv.2405.15343}
}
}

\onecolumn
\onehalfspacing

\begin{center}
\Large
\textbf{REVEAL: Robust Evolution of Vision-Language Models \\ for Explainable AI-Video Detection} \\
\vspace{0.5em}
Supplementary Material \\
\vspace{1.0em}
\end{center}

\setcounter{page}{1}

\section{More Details}
\label{supp:more_details}

\subsection{Pipeline for \vidfor Collection and Prompt Generation}
\label{supp:prompt_generation_process}

\begin{figure*}[h]
    \centering
    \includegraphics[width=1\textwidth]{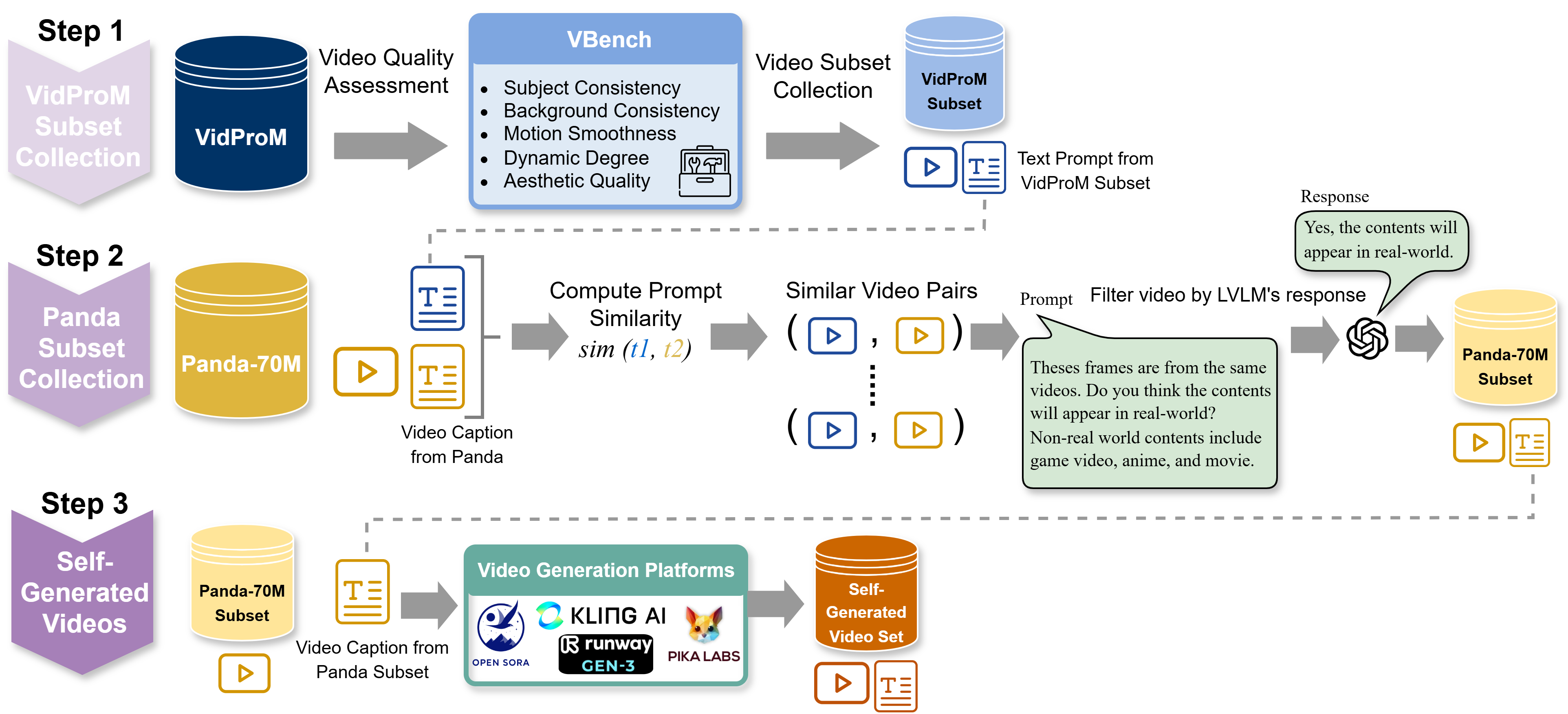}
    \caption{Dataset collection pipeline for \vidfor. Component marked with the logo$~\includegraphics[width=3mm, height=2.5mm]{figs/gpt_logo.png}$ are developed with the LVLM like GPT-4o~\cite{openai2024gpt4o}.}
    \label{fig:dataset_pipeline}
\end{figure*}

In Fig.~\ref{fig:dataset_pipeline}, we present the dataset collection pipeline for \vidfor.
 The first step is to collect fake video subset from VidProM. We use VidBench\footnote{VBench, video quality assessment tool. \url{https://github.com/Vchitect/VBench}.}, the video quality assessment tool, to filter out low-quality videos. The second step is to collect the natural video subset pairing with the VidProM subset collected from the first step. By leveraging the text prompts from the VidProM subset and video captions from Panda-70M, we compute the cosine similarity of two texts and find similar video pairs in Panda-70M. We then go through a second filtering by asking LVLM if the contents in videos are natural scenes. After finishing the second step, we collect a subset of natural videos from Panda-70M. In the third step, we use 200 real-world video captions from Panda-70M subset as the text prompts for self-generating fake videos. We use several commercial video generation platforms, including OpenSORA~\cite{openaisora}, Kling AI~\cite{klingai}, Pika Lab~\cite{pika}, and Runway Gen-3~\cite{runway3} to generate high-quality videos.

\vspace{2mm}

\lstdefinestyle{PythonStyle}{
    basicstyle=\ttfamily\small,
    stringstyle=\color{red},
    commentstyle=\color{gray},
    numbers=left,
    numberstyle=\tiny,
    stepnumber=1,
    numbersep=5pt,
    showstringspaces=false,
    breaklines=true,
    breakatwhitespace=true,
    frame=single,
    captionpos=b
}
\clearpage

\subsection{Details for Selected Explicit Knowledge}
In main paper Table~\ref{tab:ek_category}, we categorize each explicit knowledge into three video representation angles, including appearance, motion, and geometry. Here, in Table~\ref{tab:ek_category_details}, We demonstrate the understandability of LVLM on all nine explicit knowledge tools. The description of each EK tools is summarized from LVLM. In our pipeline, we select EK tools based on the reasoning ability of LVLM on them.
\label{supp:explicit-knowleged-intro}
\vspace{2mm}
\begin{table}[h]
\begin{tabular}{|l|l|l|}
\toprule
\textbf{Category}                    & \textbf{EK Name}          & \textbf{EK Description (Summarized from LVLM)}                                                                                                                                                                                                                                              \\ \midrule
\multirow{5}{*}{\textbf{Appearance}} & \textbf{Saturation}       & \begin{tabular}[c]{@{}l@{}}AI-generated videos may exhibit anomalies in color rendering. \\ Saturation estimation detects color unevenness, oversaturation, \\ or undersaturation to identify artificial elements.\end{tabular}                                               \\ \cmidrule{2-3} 
                                     & \textbf{Denoised}         & \begin{tabular}[c]{@{}l@{}}Denoising isolates unnatural noise patterns present in AI-generated \\ videos. Residual artifacts after denoising can signal synthesized \\ or forged content.\end{tabular}                                                                        \\  \cmidrule{2-3} 
                                     & \textbf{Sharpen}          & \begin{tabular}[c]{@{}l@{}}Sharpening frames emphasizes edges, making it easier to spot \\ unnatural boundaries or blending artifacts, which may indicate forgery.\end{tabular}                                                                                               \\ \cmidrule{2-3} 
                                     & \textbf{Enhance}          & \begin{tabular}[c]{@{}l@{}}Image enhancement boosts details and contrast, revealing synthetic \\ artifacts like unnatural textures or color inconsistencies.\end{tabular}                                                                                                     \\  \cmidrule{2-3} 
                                     & \textbf{Segmentation Map} & \begin{tabular}[c]{@{}l@{}}Segmentation maps identify mismatched regions in synthesized \\ content, such as areas where the object segmentation boundaries \\ do not align with real-world logic.\end{tabular}                                                                \\ \midrule
\multirow{2}{*}{\textbf{Motion}}     & \textbf{Optical Flow}     & \begin{tabular}[c]{@{}l@{}}AI-generated videos may have abnormal motion patterns, \\ such as discontinuous movements or unnatural trajectories. \\ Optical flow estimation detects whether object motion in the \\ video is smooth and adheres to physical laws.\end{tabular} \\  \cmidrule{2-3} 
                                     & \textbf{Landmark}         & \begin{tabular}[c]{@{}l@{}}In AI-generated videos, facial or body key point localization \\ may show anomalies, such as misalignment or unnatural movement. \\ Landmark estimation detects these anomalies to identify potential forgery.\end{tabular}                        \\ \midrule
\multirow{2}{*}{\textbf{Geometry}}   & \textbf{Depth Map}        & \begin{tabular}[c]{@{}l@{}}Depth information is consistent in real scenes but may exhibit anomalies \\ in AI-generated videos. Depth estimation detects issues like depth \\ dislocation and discontinuity, helping identify forged content.\end{tabular}                     \\  \cmidrule{2-3}  
                                     & \textbf{Edge}             & \begin{tabular}[c]{@{}l@{}}Synthetic videos often feature unnatural edge details, such as blurred, \\ jagged, or discontinuous object boundaries. Edge detection identifies\\  such abnormalities to pinpoint fake or synthetic elements.\end{tabular}                        \\ \bottomrule
\end{tabular}
\caption{Details for nine explicit knowledge tools}
\label{tab:ek_category_details}
\end{table}

 \clearpage
 
\subsection{Details for Explicit knowledge Preparation}
\label{supp:prompt_detail}
At the toolkit preparation step, we first call LVLMs to summarize the potential toolkit by themselves, then we further filter tools based on usefulness and capabilities. The summarization prompt shown as bellow:
\label{supp:explicit-knowleged-summarize}

\begin{tcolorbox}[colback=white!5!white,colframe=gray!2!gray,arc=2mm,fontupper=\small, left=2pt, right=2pt]
\begin{adjustbox}{minipage=[c][0.24\linewidth]{1\linewidth}}
\begin{lstlisting}[style=PythonStyle, frame=none, numbers=none]
Preparation_Prompt = """
    This is an AI-generated video detection task based on large vision-language models (LVLMs). Besides using raw frames from the video, are there any external tools that could help extract additional video information? These tools will used to facilitate LVLMs-based detection. Specifically, I'm looking for methods or tools that can generate features from the video like optical flow and sharpening. Please summarize the tool list for me.
"""
\end{lstlisting}
\end{adjustbox}
\end{tcolorbox}


We show the sample responses with three recommended features from chatGPT as follows. The keywords highlighted in blue are the useful information provided by chatGPT to facilitate our tool selection. 

\begin{tcolorbox}[colback=white!5!white,colframe=gray!2!gray,arc=2mm,fontupper=\small, left=2pt, right=2pt]
\begin{adjustbox}{minipage=[c][0.7\linewidth]{1\linewidth}}
\texttt{Certainly, for an AI video detection task based on LVLMs, leveraging external tools can significantly enhance the quality and diversity of the features extracted, which can, in turn, improve detection performance. Here are several methods and tools that can provide valuable additional features from videos:}\\

\textbf{\texttt{1. Optical Flow Extraction}} \\
\texttt{
Optical flow captures the motion between consecutive frames and can help LVLMs understand temporal dynamics. Tools such as {\color{blue}OpenCV’s Optical Flow API}(specifically {\color{blue}calcOpticalFlowFarneback} or {\color{blue}calcOpticalFlowPyrLK}) and {\color{blue}FlowNet2} can extract high-quality optical flow features that capture the movement patterns within videos.}\\

\textbf{\texttt{2. Sharpening Filters}} \\
\texttt{To enhance edges and fine details in frames, which may be beneficial for object detection or recognizing subtle transitions, consider using sharpening filters. 
{\color{blue}OpenCV’s filtering functions} ({\color{blue}cv2.filter2D} or {\color{blue}cv2.addWeighted}) can be applied to sharpen frames. Alternatively, {\color{blue}ffmpeg} offers customizable sharpen filters, which can be applied in bulk to a video.}\\

\textbf{\texttt{3. Depth Map Estimation}} \\
\texttt{Depth maps provide depth information per frame, aiding in 3D scene understanding. Tools like {\color{blue}MiDaS} (an open-source depth estimation model from Intel) and {\color{blue}DepthAI} (an AI toolkit for real-time depth estimation) can create depth maps for each frame, helping the model grasp scene structure beyond 2D features.} \\

\texttt{...... [more]}

\end{adjustbox}
\end{tcolorbox}

\clearpage

\subsection{Details for Online Adaptation on Structured Prompt}
\label{supp:prompt-adjust-process}
In this subsection, we introduce our template re-write function, which is a key component of the online adaptation process. This function ensures that the LVLMs model adapts its structured format to specific analytical needs. To achieve this, we provide a detailed system prompt consisting of five key criteria: \textit{Requirements}, \textit{Analysis Guidelines}, \textit{Constraints}, \textit{Prohibited Fields}, and \textit{Additional Notes}. These criteria collectively guide the structured prompt generation process, ensuring flexibility, precision, and relevance in adapting templates to specific tools and use cases.
\vspace{-4mm}
\paragraph{System Prompt for Template Rewrite:} \mbox{} 
\begin{tcolorbox}[colback=white!5!white,colframe=gray!2!gray,arc=2mm,fontupper=\small, left=2pt, right=2pt]
\begin{adjustbox}{minipage=[c][0.77\linewidth]{1\linewidth}}
\texttt{prompt = f"""}\\
{\color{blue}\textbf{\texttt{As a Python developer, your task is to create a new Pydantic class for analyzing video data using the tool \{tool\_name\}.}}}\\

{\color{ForestGreen}\textbf{\texttt{Requirements:}}}\\
- \texttt{The class must include the field 
'is\_ai\_generated: bool'}.\\
- \texttt{You may add any other fields you deem necessary to analyze video data with \{tool\_name\}.}\\
- \texttt{All additional fields must be of type 'str'}.\\
- \texttt{Fields should represent analysis perspectives specific to the capabilities of  \{tool\_name\}}.\\

{\color{ForestGreen}\textbf{\texttt{Analysis Guidelines:}}}\\
- \texttt{Consider the aspects of videos that \{tool\_name\} excels at analyzing.}\\
- \texttt{Reflect on patterns or anomalies that 
\{tool\_name\} might reveal.}\\
- \texttt{Emphasize high-level analysis perspectives that leverage the strengths of \{tool\_name\}.}\\

{\color{ForestGreen}\textbf{\texttt{Constraints:}}}\\
- \texttt{You may modify only one or two fields from previous class definitions at a time.}\\
- \texttt{Focus on high-level abstractions specific to the purpose of \{tool\_name\}.}\\

{\color{ForestGreen}\textbf{\texttt{Prohibited Fields:}}}\\
- \texttt{Technical parameters (e.g., frame\_rate, resolution, format, duration).}\\
- \texttt{Algorithm or implementation specifics.}\\

{\color{ForestGreen}\textbf{\texttt{Additional Notes:}}}\\
- \texttt{The total number of fields must not exceed five (5).}\\
- \texttt{There must be at least one field that differs from previous class definitions.}\\

{\color{ForestGreen}\textbf{\texttt{Previous outputs:}}} \texttt{\{history\_str\}}
\end{adjustbox}
\end{tcolorbox}

\paragraph{Template Evolution Logging} We provide the output logging for each round of template evolution. For each slot, we provide a batch of 25 real and 25 AI-generated samples as input to the LVLM. At the beginning of the first slot, we initialize the prompt template with simple key fields. In every slot, we compute the F1 score on the proposed and prior templates. If the proposed template achieves a higher F1 score, we update it; otherwise, the old template is retained. Each slot allows up to five rewrite attempts. The adaptation process terminates after several iterations. Here we set up the iteration as $4$.
\begin{tcolorbox}[colback=white!5!white,colframe=gray!2!gray,arc=2mm,fontupper=\small, left=1pt, right=1pt]
\begin{adjustbox}{minipage=[c][1.3\linewidth]{1\linewidth}}
\texttt{Starting Template Evolution with 89 Real and 89 AI-Generated Test Videos.}\\
{\color{blue}\textbf{\texttt{{--------- Slot 1/4 for edge ---------}}}}\\
\textbf{\texttt{Initial Template:}}\\
 \begin{adjustbox}{minipage=[c][0.20\linewidth]{0.5\linewidth}}
 \begin{tcolorbox}[colback=blue!10!white,,colframe=white!5!white, top=0.1pt,bottom=0.1pt]
\begin{lstlisting}[basicstyle=\footnotesize, style=PythonStyle, frame=none]
class AIAnalysisResult(BaseModel):
    is_ai_generated: bool
    raw_frame_analysis: str
    edge_analysis: str
    explanation: str
\end{lstlisting}
\end{tcolorbox}
\end{adjustbox}\\
\textbf{\texttt{Initial F1 Score: 84.94\%}}\\
\texttt{Attempt 1/5}\\
\textbf{\texttt{Proposed Template:}}\\
 \begin{adjustbox}{minipage=[c][0.22\linewidth]{0.5\linewidth}}
 \begin{tcolorbox}[colback=blue!10!white,,colframe=white!5!white, top=0.1pt,bottom=0.1pt]
\begin{lstlisting}[basicstyle=\footnotesize, style=PythonStyle, frame=none]
class NewAnalysisResult(BaseModel):
    is_ai_generated: bool
    boundary_clarity: str
    texture_consistency: str
    object_delineation: str
    spatial_anomaly_detection: str
\end{lstlisting}
\end{tcolorbox}
\end{adjustbox}\\
\textbf{\texttt{Combined F1 Score: 93.62\%}}\\
\texttt{Combined Real Success Rate: 86.36\%}\\
\texttt{Combined AI Success Rate: 100.00\%}\\
{\color{ForestGreen}$\checkmark$ \textbf{\texttt{Template improved!}}} \\
\textbf{\texttt{Slot 1 Complete}}\\
\textbf{\texttt{Best F1 Score so far: 93.62\%}}\\
{\color{blue}\textbf{\texttt{{--------- Slot 2/4 for edge ---------}}}}\\
\texttt{Evaluating previous best template...}\\
\textbf{\texttt{Previous Template F1 Score: 83.72\%}}\\
\texttt{Attempt 1/5}\\
\textbf{\texttt{Proposed Template:}}\\
 \begin{adjustbox}{minipage=[c][0.22\linewidth]{0.5\linewidth}}
 \begin{tcolorbox}[colback=blue!10!white,,colframe=white!5!white, top=0.1pt,bottom=0.1pt]
\begin{lstlisting}[basicstyle=\footnotesize, style=PythonStyle, frame=none]
class NewAnalysisResult(BaseModel):
    is_ai_generated: bool
    boundary_clarity: str
    texture_consistency: str
    object_delineation: str
    temporal_edge_coherence: str
\end{lstlisting}
\end{tcolorbox}
\end{adjustbox}\\
\textbf{\texttt{Combined F1 Score: 88.37\%}}\\
\texttt{Combined Real Success Rate: 90.91\%}\\
\texttt{Combined AI Success Rate: 86.36\%}\\
{\color{ForestGreen}$\checkmark$ \textbf{\texttt{Template improved!}}} \\
\textbf{\texttt{Slot 2 Complete}}\\
\textbf{\texttt{Best F1 Score so far: 93.62\%}}\\
\texttt{[continued...]}
\end{adjustbox}
\end{tcolorbox}

\begin{tcolorbox}[colback=white!5!white,colframe=gray!2!gray,arc=2mm,fontupper=\small, left=2pt, right=2pt]
\begin{adjustbox}{minipage=[c][0.56\linewidth]{1\linewidth}}
{\color{blue}\textbf{\texttt{{--------- Slot 3/4 for edge ---------}}}}\\
\texttt{Evaluating previous best template...}\\
\textbf{\texttt{Previous Template F1 Score: 87.50\%}}\\
{\color{BrickRed} \textbf{\texttt{Previous template performs well on new slot!}}} \\
\textbf{\texttt{Slot 3 Complete}}\\
{\color{blue}\textbf{\texttt{{--------- Slot 4/4 for edge ---------}}}}\\
\texttt{Evaluating previous best template...}\\
\textbf{\texttt{Previous Template F1 Score: 93.02\%}}\\
{\color{BrickRed} \textbf{\texttt{Previous template performs well on new slot!}}} \\
\textbf{\texttt{Slot 4 Complete}}\\
{\color{blue}\textbf{\texttt{{--------- Template Evolution Completed ---------}}}}\\
\textbf{\texttt{Final Template:}}\\
 \begin{adjustbox}{minipage=[c][0.22\linewidth]{0.5\linewidth}}
 \begin{tcolorbox}[colback=blue!10!white,,colframe=white!5!white, top=0.1pt,bottom=0.1pt]
\begin{lstlisting}[basicstyle=\footnotesize, style=PythonStyle, frame=none]
class NewAnalysisResult(BaseModel):
    is_ai_generated: bool
    boundary_clarity: str
    texture_consistency: str
    object_delineation: str
    temporal_edge_coherence: str
\end{lstlisting}
\end{tcolorbox}
\end{adjustbox}\\
\end{adjustbox}
\end{tcolorbox}

\clearpage
\section{More Results for Video-specific Tool Selection}
\label{supp:more_results} 


In Table~\ref{tab:video-specific-results}, we show the results of \reveal with video-specific tool selection, which means after selecting the toolkit for each model, when giving a test video, the model could select the tools based on its own understanding of this video, then facilitate the detection. In addition, \reveal with video-specific tool selection will further reduce the detection cost. For the Qwen-VL-Max model, the number of tools it uses per video dropped from 4 to 1.8, a decrease of 55\%; the Gemini-1.5-pro model dropped from 6 tools per video to 1.0, a decrease of 83.3\%; and the GPT-4o dropped from 3 to 2.7, a decrease of 10\%. \\
Nevertheless, the \reveal with video-specific tool selection maintains a competitive edge over the highest baseline methods. For Qwen-VL-Max, the average F1 score improves by 10. 07\% points across the eight datasets, compared to the top baseline. For Gemini-1.5-pro, the improvement is 18.25\%. And for GPT-4o, the increase is 5.93\%.

\begin{table*}[h]
\centering
\scriptsize
\setlength{\tabcolsep}{3pt} 
\renewcommand{\arraystretch}{1.5} 
\begin{tabular}{ll|cccc|ccccc}
\toprule
\multirow{2}{*}{\textbf{LVLM}} & \multirow{2}{*}{\textbf{Method}} & \multicolumn{4}{c|}{\textbf{VidForensic (VidProM)~\cite{wang2024vidprommillionscalerealpromptgallery}}} & \multicolumn{4}{c}{\textbf{VidForensic (Self-collected)}} & \multirow{2}{*}{\textbf{Avg.}} \\ \cline{3-10}
                      &                         & \textbf{Pika~\cite{pika}}   & \textbf{T2vz~\cite{khachatryan2023text2videozerotexttoimagediffusionmodels}}  & \textbf{Vc2~\cite{chen2024videocrafter2overcomingdatalimitations}}  & \textbf{Ms~\cite{wang2023modelscopetexttovideotechnicalreport}}  & \textbf{OpenSORA~\cite{opensora}} & \textbf{Gen3~\cite{runway3}} & \textbf{Kling~\cite{klingai}} & \textbf{SORA~\cite{videoworldsimulators2024}} &                       \\  
\midrule
                             
\multirow{4}{*}{\textbf{Qwen-VL-Max}~\cite{qwen2023qwenvl}} & Baseline1 (w/o SP)    & 72.50/63.09      & 75.00/67.53     & 82.00/78.57    & 76.00/69.23   & 67.50/53.24        & 62.00/40.62    & 54.50/19.47     & 58.89/39.34    & 68.55/51.24                     \\ 

                             & Baseline2 (w/o SP)        & 60.50/38.76      & 75.00/68.35    & 71.50/62.25    & 72.50/64.05   & 60.50/38.76        & 52.00/14.29    & 50.00/7.41     & 56.67/26.42    & 62.33/39.56                     \\ 
                             
                             & Baseline3 (w/o SP)        & \underline{74.00}/\underline{67.90}      & \underline{79.00}/75.58     & \underline{84.50}/\underline{83.06}    & \underline{79.50}/76.30   & 69.50/60.13        & 65.50/52.41    & 54.00/24.59     & 61.11/47.76    & 70.89/60.97                     \\ 
                             
\cline{2-11}
                             & \textbf{\reveal  (w/o SP)}         & \textbf{87.00}/\textbf{88.39}     & \textbf{81.50}/\textbf{82.63}     & \textbf{86.00}/\textbf{87.39}    & 77.00/\underline{77.45}   & \textbf{79.00}/\textbf{79.81}        & \textbf{82.50}/\textbf{83.72}    & \underline{60.00}/\underline{52.94}     & \underline{67.78}/\textbf{71.84}    & \textbf{77.60}/\textbf{76.08}                     \\
                             & \textit{w/ video-specific Sel.}            & 70.14/62.83     & 78.50/\underline{76.76}     & 82.25/81.38    & \textbf{80.17}/\textbf{78.70}   & \underline{77.25}/\underline{74.48}        & \underline{69.44}/\underline{61.53}    & \textbf{70.27}/\textbf{62.65}     & \textbf{74.02}/\underline{69.99}    & \underline{75.26}/\underline{71.04}                     \\

\midrule
\multirow{4}{*}{\textbf{Gemini-1.5-pro}~\cite{google2024gemini}} & Baseline1 (w/o SP)        & 68.33/54.32      & 71.00/59.72     & 67.00/51.47    & 75.00/67.11   & 68.50/54.68        & 64.00/44.62    & 58.00/28.81    & 58.89/41.27    & 66.34/49.83                     \\ 

                             & Baseline2 (w/o SP)        & 73.50/66.24      & 81.00/77.91     & 76.00/70.37    & \underline{85.00}/\underline{83.33}  & 71.50/62.75       & 71.50/62.75  & 59.50/37.21     & 71.11/64.86    & 72.51/58.28                    \\ 
                             
                             & Baseline3 (w/o SP)        & 64.50/45.80      & 77.00/70.51     & 71.00/59.72    & 76.50/69.68   & 64.50/45.80        & 62.00/39.68    & 52.50/11.21     & 61.11/42.62    & 66.08/51.28                     \\ 
                             
\cline{2-11}
                             & \textbf{\reveal  (w/o SP)}            & \textbf{92.00}/\textbf{91.73}     & \textbf{96.33}/\textbf{96.38}    & \textbf{95.83}/\textbf{95.87}    & \textbf{97.50}/\textbf{97.56}   & \textbf{92.17}/\textbf{91.93}        & \textbf{88.50}/\textbf{87.67}    & \underline{74.83}/\underline{68.46}     & \underline{76.67}/\textbf{78.36}    & \textbf{89.23}/\textbf{88.43}                     \\
                             & \textit{w/ video-specific Sel.}            & \underline{77.31}/\underline{71.84}     & \underline{84.00}/\underline{82.02}     & \underline{82.00}/\underline{79.25}    & 83.35/81.25   & \underline{81.50}/\underline{78.33}        & \underline{76.99}/\underline{71.29}    & \textbf{77.16}/\textbf{71.44}     & \textbf{80.09}/\underline{76.84}    & \underline{80.30}/\underline{76.53}                     \\      
\midrule
                             
\multirow{5}{*}{\textbf{GPT-4o}~\cite{openai2024gpt4o}}      & Baseline1 (w/ SP)         & 89.00/\underline{89.22}     & 90.00/90.29     & \textbf{92.50}/\underline{92.89}    & 85.00/84.69   & 82.50/81.68        & \underline{86.00}/\underline{85.86}    & 66.50/57.86     & 68.89/64.10    & 82.55/80.82                     \\  

                             & Baseline2 (w/ SP)       & 72.00/77.95      & 70.00/76.00     & 71.00/76.98    & 66.50/72.43   & 68.00/73.98        & 68.00/73.98    & 64.50/70.29   & 65.56/\underline{70.84}   & 68.20/74.06                     \\ 
                             
                             & Baseline3 (w/ SP)        & \underline{89.50}/88.66      & \underline{90.50}/\underline{90.73}     & 92.00/92.31    & 86.00/85.71  & 82.00/80.85        & 85.00/84.54    & 69.00/61.73     & 63.33/50.75   & 82.17/79.41                     \\  
                             
\cline{2-11}
                             & \textbf{\reveal  (w/ SP)}            & \textbf{93.00}/\textbf{93.46}      & \textbf{91.50}/\textbf{91.94}     & \textbf{92.50}/\textbf{92.96}   & \underline{89.00}/\underline{89.32}   & \underline{86.50}/\underline{86.57}        & \textbf{91.00}/\textbf{91.43}    &  \underline{75.50}/\underline{72.63}& \underline{68.89}/68.89    & \underline{85.99}/\underline{85.90}                     \\ 
                             & \textit{w/ video-specific Sel.}            & 84.22/83.93     & 90.00/90.65     & 90.50/91.16    & \textbf{89.67}/\textbf{90.30}   & \textbf{88.50}/\textbf{89.05}        & 83.18/82.75    & \textbf{82.46}/\textbf{81.87}     & \textbf{84.36}/\textbf{84.25}    & \textbf{86.61}/\textbf{86.75}                     \\                                   
                             
\bottomrule
\end{tabular}

\caption{Performance comparison of baselines and \reveal with and without video-specific tool selection on eight datasets. For each dataset except SORA, we mix the real dataset from Panda-70M \& AI-generated dataset together. For SORA, we mix it with 45 youtube videos that collected by ourselves. We use three representative LVLMs, including Qwen-VL-Max~\cite{qwen2023qwenvl}, Gemini-1.5-pro~\cite{google2024gemini}, and GPT-4o~\cite{openai2024gpt4o}. The results are presented as Accuracy / F1-score in each cell. Numbers in bold show the top-1 best results, and numbers with underlined show the top-2 best results.}
\label{tab:video-specific-results}
\end{table*}

\clearpage

\section{Pseudo-algorithm}
\label{supp:prompt_generation_process}
In Algo.~\ref{algo:lavid}, we provide the pseudo-algorithm for \reveal. Our detection pipeline includes two main steps (1.) EK tools selection (2.) Online adaptation for structured prompt.
\begin{algorithm}[h]
\setstretch{1.1}
\KwIn{Input Images $x$, Adaptation Set $\mathcal{X}_1$, Inference Set $\mathcal{X}_2$, Initial Prompt Template $p$, Detector $\mathcal{M}(\cdot)$, Explicit Knowledge Set $\text{EK}=\{t_1, t_2, ..., t_q\}$, Optimal Explicit Knowledge Set $\text{EK}^\star$,  Tool-Selection Metric ($S_{\text{Tool}}$,$\text{F1}_{\text{weighted}}$, $S_{\text{MP}}$), History set $\mathcal{P}_\text{history}$, Prompt Rewrite Function $\texttt{Rewrite}(\cdot)$, Detector $\mathcal{M}(\cdot)$, Number of Adaptation Iteration $\mathcal{T}$. Batch Size $B$}
\KwOut{Output prediction $\hat{y}$ from Detector $\mathcal{M}(\cdot)$.} 
\textbf{\#\#\# Adaptation} \\
    $x \sim \mathcal{X}_1, \quad \text{EK}^\star \gets \emptyset$ \Comment{Parameter onitialization}\\
    $ S_{\text{Baseline}}(x) = \alpha \cdot \text{F1}_{\text{weighted}}(x) + (1 - \alpha) \cdot S_{\text{MP}}(t=\texttt{"RGB"})$  \Comment{Compute baseline score}\\
\textbf{\#\#\# EK Tools Selection} \\
\For{$i\in\{0,...,q\}$}   
{  
       \hspace{1mm} $ S_{\text{Tool}}(t_i, x) = \alpha \cdot \text{F1}_{\text{weighted}}(t_i,x) + (1 - \alpha) \cdot S_{\text{MP}}(t_i)$  \Comment{Compute score for each tool}\\
    \If{$S_{\text{Tool}}(t_i, x) \geq S_{\text{Baseline}}(x)$} {
        $\text{EK}^\star \gets \text{EK}^\star \| t_i $ \Comment{Append tool to Optimal EK Set}
     }
     \Else{continue}
}
\textbf{\#\#\# Online adaptation for Structured  prompt} \\
$p_0 \gets p , \quad \mathcal{P}_\text{history} \gets \emptyset$  \Comment{Initialize prompt template and history set}\\
\For{$b\in\{0,...,|\mathcal{X}_2|//\text{B}\}$}{
       \hspace{1mm} $x \gets \mathcal{X}_2^{[b*B: (b+1)*B]}$   \Comment{Extract sample by batch}\\
    \For{$i\in\{0,...,\mathcal{T}\}$}{
        \hspace{1mm} $\hat{y} = \mathcal{M}(\text{EK}^\star, x, p_i)$, \quad         $\text{f1}_{score} = \text{F1}(\hat{y}, y)$  \Comment{Compute score for current prompt}\\
        $p'_i \gets \texttt{Rewrite}(p_i, \mathcal{P}_\text{history})$ \Comment{Rewrite prompt}\\
        $\hat{y}' = \mathcal{M}(\text{EK}^\star, x, p'_i)$, \quad         $\text{f1}'_{score} = \text{F1}(\hat{y}', y)$  \Comment{Compute score for rewritten prompt}\\
        \If{$\text{f1}^{'}_{score} \geq \text{f1}_{score}$}{
         $ \mathcal{P}_\text{history} \gets  \mathcal{P}_\text{history} \| (p'_i, \text{f1}'_{score})$  \Comment{Append rewritten prompt to history set}\\
        }
        \Else{
         $ \mathcal{P}_\text{history} \gets  \mathcal{P}_\text{history} \| (p_i, \text{f1}_{score})$ \Comment{Append current prompt to history set}\\ 
        }
    \hspace{1mm} $p^\star \gets p_{\mathcal{T}}$ \\
    $\hat{y}^\star =\bigcup_{i=1}^{n}\mathcal{M}(\text{EK}^\star_i, x, p^\star)$ \Comment{Union the prediction for batch sample} 
    }
}
    

\caption{\textbf{Pseudo-algorithm for \reveal detection pipeline}}
\label{algo:lavid}
\end{algorithm}%

\clearpage

\section{Sample Visualization} 
\label{supp:sample-visualization}
\begin{figure}[h!]
    \centering
    \includegraphics[width=.999\linewidth]{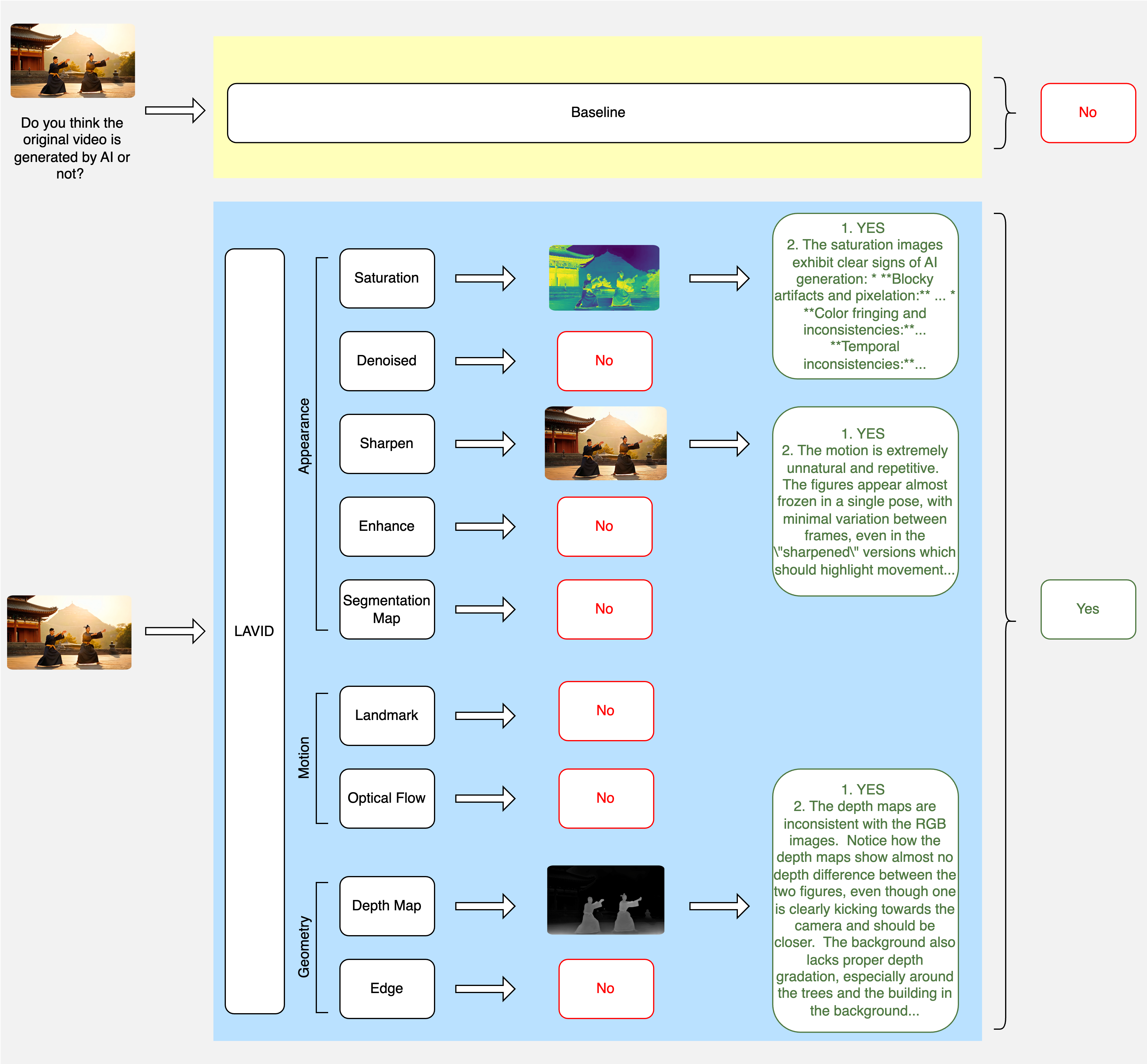}
    \caption{Sample Visualization}
    \label{fig:data_visualization}
\end{figure}

In Fig.~\ref{fig:data_visualization}, we provide a visualization of diffusion-generated video detection through our agentic framework LAVID and a baseline model. LAVID analyzes the video using EK tools and utilizes the results of the analysis to make robust detection decisions while the baseline model simply outputs a "yes" or "no" when prompted "do you think the original video is generated by AI or not?".

\end{document}